\documentclass{article} %
\usepackage{iclr2027_conference,times}

\usepackage{amsmath,amsfonts,bm}

\def\eqref#1{equation~\ref{#1}}

\def\1{\bm{1}}

\DeclareMathAlphabet{\mathsfit}{\encodingdefault}{\sfdefault}{m}{sl}
\SetMathAlphabet{\mathsfit}{bold}{\encodingdefault}{\sfdefault}{bx}{n}

\usepackage{hyperref}
\usepackage{url}
\usepackage{float}
\usepackage{amssymb}

\usepackage{subcaption}
\usepackage{float}

\usepackage{tabularray}
\UseTblrLibrary{booktabs}
\usepackage{rotating}
\usepackage{multirow} 
\usepackage{amsmath,amsfonts,bm}
\usepackage{tabularx}
\usepackage{pdflscape}
\usepackage{ulem}
\usepackage{booktabs}
\usepackage{xcolor}
\usepackage{wrapfig}
\usepackage{caption}
\usepackage{graphicx}
\usepackage{wrapfig}
\usepackage{tikz}
\usetikzlibrary{arrows.meta}
\usepackage{array}

\usepackage{placeins}
\usepackage{dblfloatfix}

\usepackage{pgfplots}
\pgfplotsset{compat=1.18}

\usepackage{adjustbox}  %
\usepackage{enumitem}
\usepackage{soul}
\usepackage{soulpos}
\usepackage[most]{tcolorbox}
\usepackage{caption}
\DeclareCaptionFont{tinyfig}{\fontsize{6}{7}\selectfont}

\title{Learning to Design SR Solvers: Harness Optimization for Symbolic Regression}

\author{Qian Li$^{1,2}$, Yuxiao Hu$^{2,3}$, Qihao Duan$^{4}$, Jinyue Yan$^{3}$, Yuntian Chen$^{2}$\thanks{Corresponding author} \\
$^{1}$Shanghai Jiao Tong University, Shanghai, China\\
$^{2}$Eastern Institute of Technology, Ningbo, China\\
$^3$The Hong Kong Polytechnic University, Hong Kong, China\\
$^4$Epigenetics Laboratory, Max Planck Institute for Heart and Lung Research\\
\texttt{qianl01205@sjtu.edu.cn}, \texttt{yuxiao.hu@connect.polyu.hk}\\
\And
}

\iclrfinalcopy %

\definecolor{promptfill}{rgb}{0.95,0.97,1.0}
\definecolor{promptstroke}{rgb}{0.2,0.4,0.7}

\begin{document}

\maketitle
\lhead{Preprint}

\begin{abstract}
Symbolic regression (SR) seeks explicit equations that explain observed data, and recent methods increasingly employ large language models (LLMs) and agents for equation discovery. However, these systems typically rely on manually designed workflows and prompts. We argue that the SR agent harness should itself be an optimization target: its design shapes search priors and how numerical feedback guides subsequent equation refinement. We therefore propose \textbf{HarnessSR}, a framework for jointly optimizing prompts and workflows to construct reusable SR agents. During both harness optimization and agent execution, HarnessSR supports either PySR or an LLM-based equation generator through tool-specific interfaces, allowing the framework to optimize search guidance for different equation-discovery mechanisms.
HarnessSR first optimizes role prompts through equation-discovery feedback in basic workflows, then jointly searches prompt assignments and workflows with brief adaptation to align the two. Finally, it refines role instructions through reflection on residual patterns, candidate equation structures, and search outcomes, improving how these signals guide subsequent equation search. On LSR-Synth in LLM-SRBench, HarnessSR achieves a mean ID $\mathrm{Acc}_{0.01}$ of 61.33\% across three backbones, a 45.0\% relative improvement over the next-best baseline.
\end{abstract}

\section{Introduction}
Symbolic regression (SR) seeks explicit mathematical relationships that explain observed data~\citep{wang2023scientific,zhang2025artificial,karniadakis2021physics,krenn2022scientific,udrescu2020ai,udrescu2020ai2}. The resulting equations can be inspected, interpreted, and manipulated, and related to conservation laws, dimensional constraints, and domain theories. SR thus goes beyond numerical prediction to turn observations into testable scientific hypotheses, supporting equation discovery across physics, biology, materials science, and other domains.

Recent advances in large language models (LLMs) and agents have introduced a new paradigm for SR. These methods combine the scientific knowledge and reasoning capabilities of LLMs with numerical search, evaluation, and feedback to iteratively propose, test, and revise equation hypotheses~\citep{xia2025sr,llmsr,yang2026think,zhao2026sr}. Central to these systems is the \textit{solver harness}: the executable layer of role instructions, tool interfaces, and control logic that governs interactions between the LLM, data, and numerical tools. The harness governs how scientific knowledge informs candidate proposals and how numerical feedback guides decisions to retain, revise, or replace them. These decisions shape whether subsequent computation explores alternative functional forms, addresses structural errors, or repeats attempts, thereby affecting the quality of equations recovered within a given budget.

However, recent SR solver largely focus on designing solver harnesses by hand, rather than treating the solver harness itself as an object of systematic optimization. Recent research shows that, even with the base model fixed, harness choices can affect both task performance and inference resource use~\citep{lee2026meta,zhou2026multi}. Solver harness optimization extends beyond adding new SR roles. Within a fixed workflow, different instructions and few-shot demonstrations can shift the model’s output distribution. Conversely, with SR roles and prompts held fixed, changes in workflow topology—including invocation frequency and parallelism—can alter information flow, computational allocation, and final outcomes. These choices influence how data analysis informs equation search, how fitting and residual feedback guide candidate revision, and how finite computation is allocated between exploration and refinement. \textbf{The SR agent harness is therefore not merely an execution scaffold, but a design variable to be optimized for equation discovery.} This motivates us to develop a method for automatically optimizing the harness for reusable SR agent, aiming to improve both the quality and efficiency of equation discovery.

To this end, we introduce \textbf{HarnessSR}, a framework for optimizing reusable SR solver harnesses. We formulate harness design as the joint optimization of role instructions, demonstrations, and workflow organization to obtain an SR agent harness capable of discovering higher-quality equations. To address the strong coupling among these components and tailor their behavior to SR, we adopt a three-stage procedure. First, we optimize role prompts within basic workflows to establish strong starting points for joint search. Second, we jointly optimize workflows and prompt assignments, with brief adaptation to accommodate their interdependence. Third, we use a reflection model to diagnose failures in SR execution traces and refine role instructions, enabling the harness to translate residual patterns, candidate equation structures, and search outcomes into targeted search guidance.
This guidance is applied through tool-specific interfaces that translate agent-generated search plans into parameters and inputs for two selectable base SR tools (we integrates PySR and an LLM-based equation generator as agent tools). These interfaces allow agents to steer the chosen tool during both harness optimization and deployment, while keeping its underlying search algorithm fixed.
The top-performing harness is then frozen and reused across new SR tasks, with model weights and the underlying search algorithms held fixed. Across three backbones on LSR-Synth, HarnessSR achieves a mean ID $\mathrm{Acc}_{0.01}$ of 61.33\%, a 45.0\% relative improvement over the next-best baseline.

Our main contributions are: (1) We highlight that the solver agent harness can also be regarded as an optimization target for SR task. Its design shapes the LLM’s reasoning process, influencing the priors that guide equation search, how search feedback is interpreted and used, and ultimately the quality of the discovered equations. (2) We propose HarnessSR, an SR agent optimization framework that jointly optimizes prompts and workflows and uses reflection on SR execution traces to adapt harness behavior to equation discovery. (3) We demonstrate a 45.0\% relative improvement in mean ID $\mathrm{Acc}_{0.01}$ over the next-best baseline across three backbones on LSR-Synth.

\section{Related Work}
\label{gen_inst}
\textbf{Symbolic Regression.\quad}
Classical SR methods include genetic programming, which evolves expressions through crossover, and mutation~\citep{gp,gplearn,pawlak2018competent,
moraglio2012geometric,de2021interaction}. PySR combines evolutionary search with expression simplification and constant optimization~\citep{pysr}. Neural-guided methods such as DSR learn a distribution over expressions using a recurrent neural network and guide candidate generation through risk-seeking policy gradients~\citep{dsr,zhang2025rag,gensr,meidani2023snip,e2e,transr,shojaee2023transformer,yu2025discover}.

LLM-based methods leverage the generative capacity, prior knowledge, and reasoning capabilities of LLMs to guide equation search. LLM-SR and SGA couple LLM-generated structural hypotheses with numerical parameter optimization~\citep{llmsr,ma2024llm}. Beyond candidate generation, LaSR introduces a learned concept library to guide evolutionary search~\citep{lasr}, while RESTART uses residual feedback to guide an LLM in revising candidate equation structures~\citep{li2026robust}. Deliberate Evolution guides LLM-based equation generation with adaptive search operators, structural diagnostics, and reflective memory~\citep{pang2026deliberate}. 

Agentic methods extend this paradigm through tool use and feedback-driven interaction. SR-Scientist autonomously invokes data-analysis and equation-evaluation tools to refine hypotheses~\citep{xia2025sr}. KeplerAgent extracts physical properties, including symmetries, to configure numerical SR engines~\cite{yang2026think}. A-SR introduces hierarchical coordination to adapt role selection, memory use, and feedback routing during search~\citep{zhao2026sr}.

Despite increasingly autonomous execution, the overarching harness in representative SR systems remains largely hand-designed. For example, SR-Scientist selects tools autonomously, but does so within predefined tool interfaces, instruction templates, and outer-loop control rules. Such autonomy adapts the search within a given harness without optimizing the harness itself. Yet these design choices govern how hypotheses are proposed and how evidence guides subsequent search, making them integral to the equation-discovery strategy. Therefore, we focus on optimizing the solver harness under a limited budget to obtain stronger, reusable SR solvers.

\textbf{Self-Evolving Agents.}\quad
Self-evolution and automated agent design extend optimization from individual outputs to the procedures that produce them. MIPRO optimizes instructions and demonstrations using end-to-end task feedback~\citep{opsahl2024optimizing}. MASS optimizes prompts for individual building blocks, selects how these blocks are composed, and then refines their prompts within the selected workflow~\citep{zhou2026multi}, while Meta-Harness searches executable code surrounding a fixed language model using execution feedback~\citep{lee2026meta}. However, existing methods have devoted limited attention to harness optimization for automated scientific discovery. In our paper, we show that optimizing the solver harness can substantially improve both the quality and efficiency of scientific discovery.

\begin{figure}[H]
 \includegraphics[width=\textwidth]{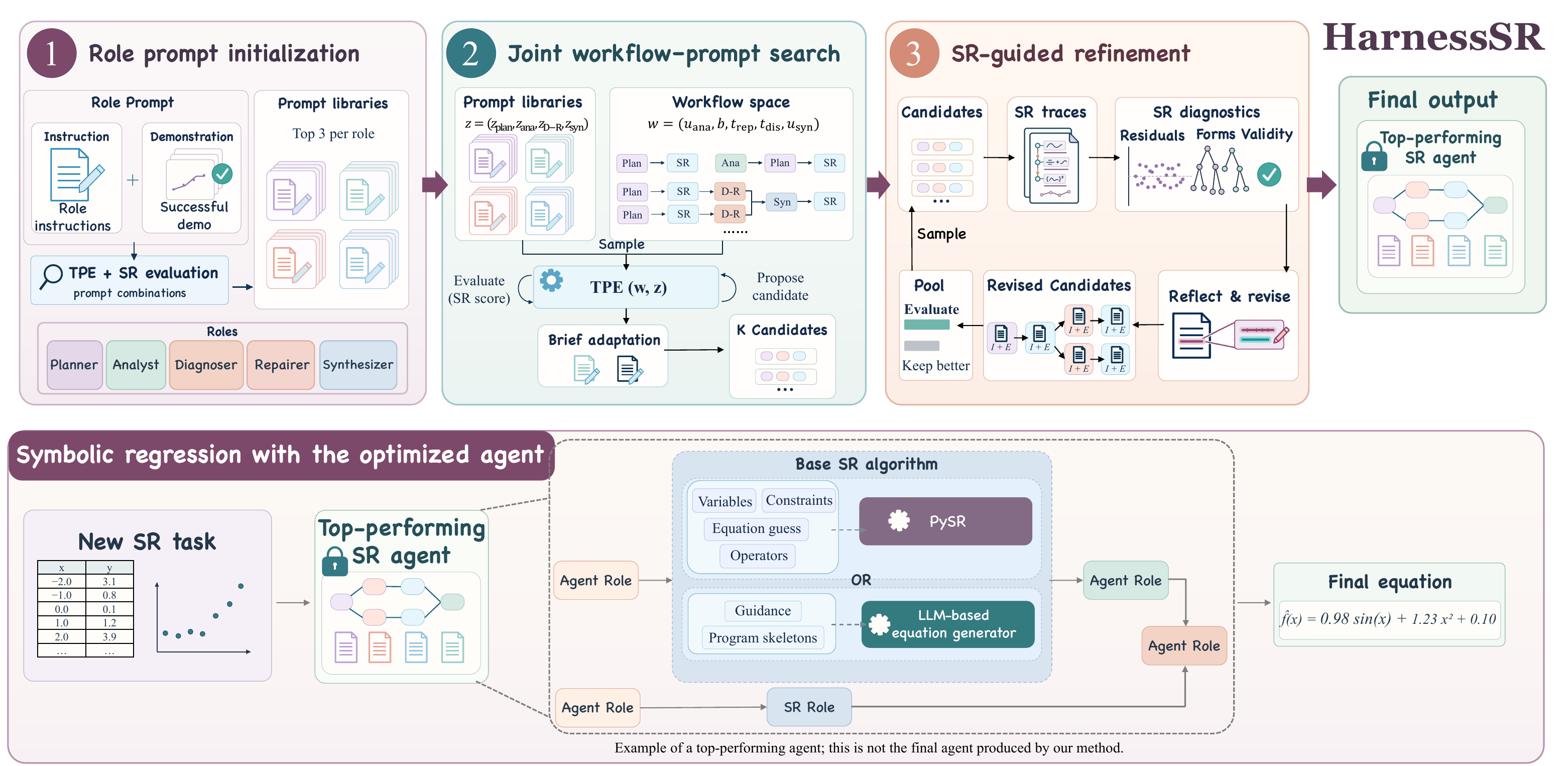}
  \caption{Overview of HarnessSR. It initializes role prompts, jointly optimizes workflows and prompt assignments, and refines instructions with SR-specific feedback. The top-performing agent harness is then frozen and reused on new tasks.
The agent workflow shown below is illustrative.}
\vspace{-10pt}
  \label{overview}
\end{figure}

\section{Method}\label{sec:method}
\label{headings}
Unlike prior SR methods that manually design a new solver, we optimize the solver harness itself. For an SR agent, HarnessSR jointly searches role prompts, activation, parallelism, and iteration. To coordinate these coupled components for equation discovery, we adopt a three-stage procedure. The resulting top-performing SR agent is reusable and better coordinates scientific reasoning with numerical search under limited compute.

\subsection{Problem Formulation}
Given an SR task \(q\), typically comprising scientific context, variable descriptions, and observed data, an SR agent \(S\) seeks an explicit equation explaining the observations. With the base language model \(M\) fixed, \(S\) is instantiated from its harness \(\mathcal H\) and returns \(\hat f_q\):
\[
S=\operatorname{Instantiate}(M,\mathcal H),
\qquad
\hat f_q=S(q).
\]
We aim to optimize SR agents by optimizing their harnesses without changing the LLM weights.

We define \(\mathcal H=(\mathcal R,w,\mathcal F)\), where \(\mathcal R\) specifies the typical roles in SR agent and their prompts:
\[
\mathcal R=\left\{(r,P_r)\;\middle|\;r\in\{
\mathrm{Planner},\mathrm{Analyst},\mathrm{Diagnoser},
\mathrm{Repairer},\mathrm{Synthesizer}\}\right\}.
\]
These roles respectively construct equation search plans, extract evidence of observation, diagnose equation deficiencies, revise plans based on diagnoses, and integrate findings across branches. The \(w\) controls the workflow structure. \(\mathcal F\) contains fixed tools and components shared by all candidates, including a base SR tool such as PySR~\citep{pysr} and an LLM-based equation generator, candidate management, and the final equation selector.

For each role \(r\), the full prompt includes runtime context, such as task information, search history, and feedback from other roles, together with two optimizable components $P_r=(I_r,E_r)$: an instruction $I_r$ and a demonstration set $E_r$. The instruction \(I_r\) specifies the role’s objectives, reasoning requirements, and action principles. The demonstration set \(E_r\) illustrates the role’s task through input–output examples and may be empty. These components can alter the LLM's reasoning strategies and outputs even with roles and workflows fixed.

We define the workflow using five discrete variables:
\[w=\left(u_{\mathrm{ana}},b,t_{\mathrm{rep}},t_{\mathrm{dis}},u_{\mathrm{syn}}\right).
\]
These variables jointly control role activation, parallel search, and iterative refinement, as summarized in Tab.~\ref{tab:workflow_space}.
\begin{table}[H]
\centering
\captionsetup{font=scriptsize,skip=2pt}
\vspace{-8pt}
\caption{Effects of workflow parameters. PySR as an example of base SR tools.}
\label{tab:workflow_space}
\fontsize{5.6}{6}\selectfont
\setlength{\tabcolsep}{4pt}
\renewcommand{\arraystretch}{1.05}
\begin{tabularx}{\textwidth}{
    @{} l l >{\raggedright\arraybackslash}X @{}
}
\toprule
Parameter & Values & Effect on the workflow \\
\midrule
$u_{\mathrm{ana}}$
& $\{0,1\}$
& Activates \textit{Analyst} to examines observational evidence before initial planning. \\
$b$
& $\{1,2,3,4,5\}$
& Number of parallel search branches, each executing \textit{Planner} $\rightarrow$ PySR, or \textit{Planner} $\rightarrow$$\cdots$$\rightarrow$ PySR \\
$t_{\mathrm{rep}}$
& $\{0,1,2,3\}$
& Number of search-plan refinement rounds in each branch, following \textit{Diagnoser} $\rightarrow$ \textit{Repairer} $\rightarrow$ PySR. \\
$t_{\mathrm{dis}}$
& $\{0,1,2\}$
& Number of cross-branch discussion rounds; in each round, each branch uses the shared results to update its own search plan and run PySR.\\
$u_{\mathrm{syn}}$
& $\{0,1\}$
& Activates \textit{Synthesizer} to integrate branch findings into a unified plan, followed by one additional PySR run.\\
\bottomrule
\vspace{-18pt}
\end{tabularx}
\end{table}
The task-initiating \textit{Planner} is always active and needs no control parameter. The \textit{Diagnoser}--\textit{Repairer} pair is active when \(t_{\mathrm{rep}}>0\). Single-branch workflows omit cross-branch replanning and synthesis, requiring \(t_{\mathrm{dis}}=u_{\mathrm{syn}}=0\) when \(b=1\). Sec.~\ref{3.2} details the execution order and information flow.

We jointly optimize role instructions, demonstrations, and workflow organization on a training set \(\mathcal D\) to maximize average equation quality under workflow-validity and execution-budget constraints:
\[
(\mathcal R^\star,w^\star)\in\underset{\substack{\mathcal R,\,w\\w\in\mathcal W_B}}{\arg\max}\;J_{\mathcal D}(\mathcal H),
\ \text{where}\
J_{\mathcal D}(\mathcal H)
=
\frac{1}{|\mathcal D|}
\sum_{q\in\mathcal D}
Q\!\left(S(q),q\right).
\]

Here, \(\mathcal W_B\) contains workflows satisfying role dependencies and a per-task execution budget \(B\). Each candidate evaluation executes the complete solver and scores its final equation using the fixed quality score \(Q\). The stage-specific optimization sets \(\mathcal{D}=\cup\{\mathcal D_i\}_{i=1}^{3}\) and selection set \(\mathcal D_{\rm eva}\) are mutually disjoint random subsets of the public training set released by~\cite{xia2025sr}. Each \(\mathcal D_i\) is used for optimization in Stage \(i\), while \(\mathcal D_{\rm eva}\) is used for candidate selection during harness optimization. The three-stage optimization of HarnessSR approximately solves this objective to obtain the top-performing SR solver agent, and then applies it to new SR tasks.

\subsection{Execution within an SR Agent Harness}\label{3.2}
Given task \(q\), the harness optionally invokes the \textit{Analyst} to examine observations, then initializes \(b\) branches. Each branch executes a \textit{Planner}-generated plan using a base SR tool, followed by \(t_{\mathrm{rep}}\) \textit{Diagnoser}--\textit{Repairer} refinement rounds. In each of the \(t_{\mathrm{dis}}\) cross-branch replanning rounds, each \textit{Planner} uses a shared snapshot of preceding results to initiate another search. An optional \textit{Synthesizer} produces one additional plan. Each role shares its prompt \(P_r\) across branches and rounds, while runtime context changes.

We integrate two selectable base SR algorithms as agent tools: PySR or an LLM-based equation generator, through a fixed, tool-specific adapter. For PySR, role outputs specify variable subsets, operator sets (\texttt{unary\_operators}, \texttt{binary\_operators}), structural restrictions (\texttt{constraints}, \texttt{nested\_constraints}), and expression seeds (\texttt{guesses}) to constrain and seed evolutionary search. For the LLM-based generator, the adapter incorporates search guidance and structural hypotheses into the generation prompt and converts expression guesses into initial parameterized program skeletons. The generator retains its iterative equation-program generation, numerical parameter fitting, and evaluation procedures. For both tools, harness updates guide search through tool inputs while leaving the underlying algorithms unchanged.

Harness updates modify the parameters passed to the base SR algorithm. The shared components \(\mathcal F\) collect valid candidates and select the best equation \(\hat f_q\), which is scored by an external evaluator as \(Q\). Subsequent stages optimize prompts and workflows through this execution pipeline.

\subsection{Stage 1: Role-Level Prompt Optimization}
Stage 1 needs to identify a strong base prompt set for each single role, providing a starting point for the joint optimization of prompts and workflow in Stage 2. We optimize instructions $I_r$ and demonstrations $E_r$ of each role $r$ within simple workflows, retaining three high-scoring prompts per role for subsequent search and adaptation. 

\textbf{Optimization of Planner' base prompts.} We first fix the base workflow \(w=(0,1,0,0,0)\), which structure is detailed in Tab.~\ref{tab:role_initialization}, and optimize the \textit{Planner}'s instructions $I_{\rm Planner}$ and demonstrations $E_{\rm Planner}$. By running the solver agent on \(\mathcal D_1\), we extract input–output demonstrations  from successful solver trajectories and use them, together with task information, to generate candidate instructions. Tree-structured Parzen estimator (TPE)~\citep{bergstra2011algorithms} searches over instruction--demonstration combinations, evaluating each through complete solver agent executions on \(\mathcal D_{\rm eva}\). We retain the three highest-scoring combinations as the Planner's base prompt packages for subsequent workflow search.

\begin{wraptable}{r}{0.64\linewidth}
\vspace{-15pt}
\centering
\captionsetup{font=scriptsize,skip=2pt}
\caption{Local workflows for prompt initialization. PySR as an example of base SR tools.}
\label{tab:role_initialization}
\fontsize{5.8}{6.2}\selectfont
\setlength{\tabcolsep}{2pt}
\renewcommand{\arraystretch}{0.85}
\setlength{\aboverulesep}{1pt}
\setlength{\belowrulesep}{1pt}

\begin{tabularx}{\linewidth}{
    @{} l >{\raggedright\arraybackslash}X c @{}
}
\toprule
Target & Initialization workflow & $w$ \\
\midrule
Planner
& Planner $\to$ PySR $\to$ Selection
& (0,1,0,0,0) \\
Analyst
& Analyst $\to$ Planner $\to$ PySR $\to$ Selection
& (1,1,0,0,0) \\
Diag--Re
& Planner $\to$ PySR $\to$ Diagnoser $\to$ Repairer
$\to$ PySR $\to$ Selection
& (0,1,1,0,0) \\
Synthesizer
& Two parallel [Planner $\to$ PySR] branches
$\to$ Synthesizer $\to$ PySR $\to$ Selection
& (0,2,0,0,1) \\
\bottomrule
\end{tabularx}
\vspace{-8pt}
\end{wraptable}

\textbf{Optimization of other roles' base prompts.} With the \textit{Planner}'s base prompt fixed, we optimize each remaining role within its corresponding workflow in Tab. \ref{tab:role_initialization}, using the same prompt optimization procedure. Only the target roles' instructions and demonstrations are updated; all other components remain fixed. \textit{Diagnoser} and \textit{Repairer} prompts are optimized and retained as paired configurations.

\textbf{Prompt libraries.} We remove duplicate and invalid versions, and retain the three highest-scoring prompts per role. Diagnoser and Repairer prompts remain paired, yielding four libraries:
\[
\mathcal L_r=\{P_r^{(1)},P_r^{(2)},P_r^{(3)}\},
\quad
r\in\{\mathrm{Planner},\mathrm{Analyst},
\mathrm{DiagRepair},\mathrm{Synthesizer}\}.
\]
Each package contains one \(P_r=(I_r,E_r)\), or a paired diagnosis–repair configuration.

These prompts are initial candidates rather than final configurations. Stage 2 jointly optimizes both workflow and prompts, while Stage 3 further refines the instructions using SR execution feedback for task adaptation.

\subsection{Stage 2: Joint Workflow and Prompt Optimization}
\label{3.4}
Stage 2 jointly searches workflows and prompt assignments, then briefly adapts the selected prompts to their workflows before choosing candidates for Stage 3. 

\begin{wrapfigure}{r}{0.34\textwidth}
    \centering
    \vspace{-17pt}
    \includegraphics[width=\linewidth]{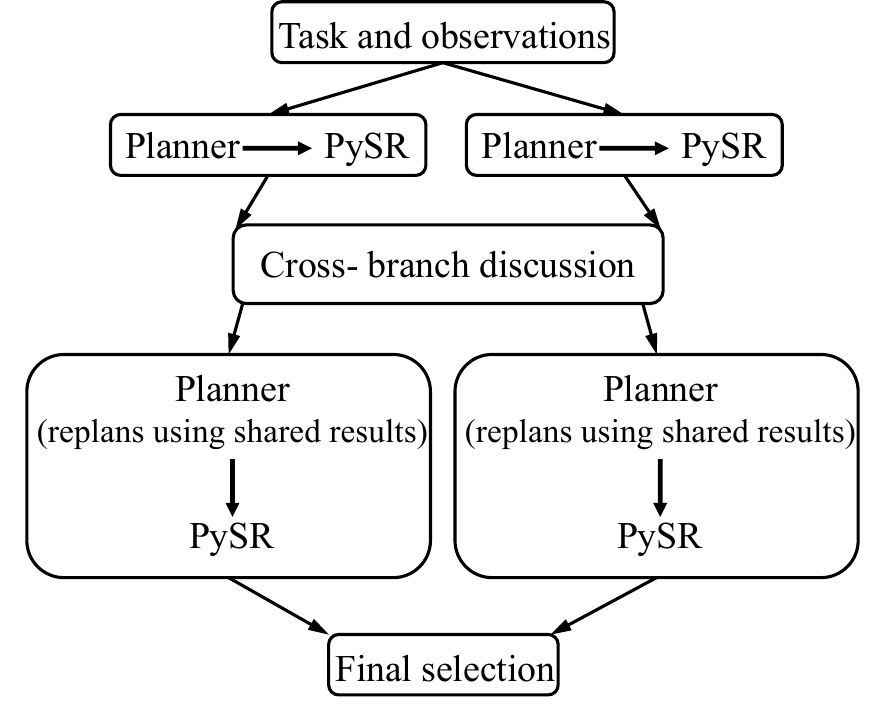}
    \captionsetup{font=tiny,skip=3pt}
    \caption{2-branch search with 1 cross-branch
    discussion round: $w$=(0,2,0,1,0). PySR as an example.}
    \vspace{-20pt}
    \label{fig:cross_branch_discussion}
\end{wrapfigure}
\textbf{Candidate agent initialization.} We initialize each agent candidate by sampling a workflow and then assigning prompts to it.

Firstly, we construct workflow space $\mathcal W_B$ by enumerating the parameter combinations in Tab.~\ref{tab:workflow_space} and removing those that violate role dependencies or execution-budget constraints. We select $n$ distinct workflows from $\mathcal W_B$, which include six basic workflows (the four listed in Tab. \ref{tab:role_initialization}, two-branch independent search $(0,2,0,0,0)$, and cross-branch replanning $(0,2,0,1,0)$, illustrated in Fig.~\ref{fig:cross_branch_discussion}). The remaining workflows are sampled uniformly without replacement from $\mathcal W_B$.

Secondly, for each sampled workflow, we assign a prompt to every role by probabilistic sampling from the corresponding Stage 1 library: higher-scoring packages are more likely to be selected, while lower-scoring ones remain available for exploration. We repeat this assignment twice for each workflow, resampling duplicate assignments, to obtain $2n$ initial harnesses. 

We represent each candidate agent by $(w,z)$, where $w$ specifies the workflow and $z=\left(z_{\mathrm{Planner}},z_{\mathrm{Analyst}},z_{\mathrm{DiagRepair}},z_{\mathrm{Synthesizer}}\right)$ specifies its prompt assignment. Each $z_r\in\{1,2,3\}$ selects package $P_r^{(z_r)}$ from $\mathcal L_r$, determining the corresponding role prompts $P_r$ in $\mathcal R$. Only indices for active role groups are used; \textit{Diagnoser} and \textit{Repairer} share one package index.

\textbf{Joint search with TPE.}
We use the TPE~\citep{bergstra2011algorithms} to optimize $(w,z)$: the five workflow parameters in $w$ and up to four active prompt-package indices in $z$. 
We first evaluate the $2n$ initial harnesses on $\mathcal D_2$ and initialize TPE
with their configurations and scores. At each iteration, TPE divides evaluated configurations into high-scoring candidates and the remainder, models their parameter distributions, and proposes a new $(w,z)$ favored by the high-scoring distribution relative to the other. The resulting solver is evaluated on the same panel, and its score updates the search history. This loop can change both workflow structure and prompt assignment across the full valid space. Therefore, during TPE search, the prompts obtained in Stage 1 remain fixed and are only reassigned across different workflows. Invalid or previously evaluated configurations are excluded.

\textbf{Brief prompt adaptation for top candidates.}
TPE can only search for better prompt–workflow combinations within a fixed space and cannot generate new prompts. Because prompts and workflows are tightly coupled, Stage 1 prompts may become suboptimal for the workflows found in Stage 2 and bias their evaluation. We therefore briefly refine the each role's instruction $I_r$ of promising candidates before allocating the larger Stage 3 optimization budget.

We select the best $K$ candidate agents with different workflows based on joint search and fix their workflows and demonstrations. For every role of each top candidate, we ask an LLM to propose a small set of instruction variants informed by role input, output, and execution feedback. TPE then searches for combinations of these variants, evaluating each combination through complete solver executions on a small subset of $\mathcal D_2$. Finally, we retain the best instruction combination for each top candidate, effectively fine-tuning its prompts.

\subsection{Stage 3: SR-Guided Instruction Refinement}
\label{3.5}
Brief adaptation improves prompt–workflow compatibility but cannot fully address persistent failures in equation search. Stage 3 therefore refines the \(K\) Stage 2 candidates using SR-specific execution feedback. For each candidate, we fix the workflow \(w\), demonstrations \(E_r\), and shared components \(\mathcal F\), and optimize only the role instructions \(I_r\) to better adapt them to SR tasks.

\textbf{SR diagnostic feedback.}
Each workflow maintains a separate candidate pool $\mathcal C_w$,
initialized with its adapted harness from Stage 2. At each iteration,
we uniformly sample a parent $\mathcal H_w$ from the pool and execute it
on $\mathcal D_3$, recording every role's input and output, search plans, candidate equations, and observational residuals.

An external diagnostic LLM examines these traces and reports identified
problems, supporting evidence, and affected roles. It checks and constructs four types of diagnostic feedback:
(i) evidence-to-action alignment, checking whether residual
patterns and diagnoses lead to concrete search-plan changes;
(ii) structural exploration, examining whether branches and
rounds explore distinct equation forms and improve candidate quality;
(iii) constraint justification, identifying variable, operator,
or structural restrictions unsupported by available evidence; and
(iv) expression quality, examining fit, complexity, and numerical
validity to inform subsequent planning. 

This step diagnoses behavior without rewriting instructions or accessing reference equations, ensuring that the diagnostics are not tailored to specific training problems.

\textbf{Feedback-guided instruction revision.}
The reflection model~\citep{agrawal2026gepa} receives the active roles' current instructions and the diagnostic feedback, proposes evidence-supported changes, and produces revised instructions for every roles of candidate workflow. 
The resulting prompts form a child harness $\mathcal H_w'=(\mathcal R'=\{(r,P'_r)\},w,\mathcal F)$. 

\textbf{Determining the Final SR Harness.}
Parent and child are evaluated on $\mathcal D_{\rm eva}$. The child enters $\mathcal C_w$
only if
\[
J_{\mathcal D_{\rm eva}}(\mathcal H')>J_{\mathcal D_{\rm eva}}(\mathcal H).
\]
Otherwise, it is discarded. We repeat this process until the optimization budget is exhausted. 

We evaluate the candidates in each pool on \(\mathcal D_3\) and retain the highest-scoring harness for each of the \(K\) workflows. We then evaluate these \(K\) harnesses on \(\mathcal D_3 \cup \mathcal D_{\mathrm{eva}}\) and select the top-performing one as the final SR agent harness.

\begingroup
\normalem
\section{Experiments}\label{sec:experiments}
\raggedbottom
\captionsetup{font=small,skip=4pt}
We evaluate HarnessSR's accuracy and efficiency, compare harness optimization with model-weight post-training, and analyze the contributions of optimization stages, workflow search, and diagnostic feedback. The appendices provide experimental configurations and role instructions (\ref{app:settings}--\ref{app:harness} and \ref{app:prompts}), alongside optimization costs ( \ref{app:cost}), symbolic accuracy (\ref{app:sa}), alternative SR backend comparisons (\ref{app:harness}), out-of-distribution results (\ref{app:ood}), equation-recovery cases (\ref{app:cases}), and training-set-size analysis (\ref{app:training-size}).

\begin{table}[H]
\centering
\caption{ID accuracy (\%) and mean solving time (min/task) on LSR-Synth. Bold indicates the best result in each column.}
\label{tab:main}
\begingroup
\fontsize{8.3}{9.6}\selectfont
\setlength{\tabcolsep}{1.35pt}
\renewcommand{\arraystretch}{0.92}
\setlength{\aboverulesep}{0.8pt}
\setlength{\belowrulesep}{0.8pt}
\newcommand{\mainmetric}[1]{$\mathrm{Acc}_{#1}$}
\begin{tabular*}{\textwidth}{@{\extracolsep{\fill}}l*{11}{c}@{}}
\toprule
\multirow{2}{*}{\textbf{Method}} & \multicolumn{2}{c}{\textbf{Overall}} & \multicolumn{2}{c}{\textbf{Physics}} & \multicolumn{2}{c}{\textbf{Chemistry}} & \multicolumn{2}{c}{\textbf{Biology}} & \multicolumn{2}{c}{\textbf{Materials}} & \multirow{2}{*}{\shortstack{\textbf{Time} $\downarrow$\\[-1pt]min/task}} \\
\cmidrule(lr){2-3}\cmidrule(lr){4-5}\cmidrule(lr){6-7}\cmidrule(lr){8-9}\cmidrule(lr){10-11}
& \mainmetric{0.01} & \mainmetric{0.001} & \mainmetric{0.01} & \mainmetric{0.001} & \mainmetric{0.01} & \mainmetric{0.001} & \mainmetric{0.01} & \mainmetric{0.001} & \mainmetric{0.01} & \mainmetric{0.001} & \\
\midrule
\multicolumn{12}{c}{\textit{Non-LLM methods}}\\
\midrule
GPLearn & 0.00 & 0.00 & 0.00 & 0.00 & 0.00 & 0.00 & 0.00 & 0.00 & 0.00 & 0.00 & 3.15 \\
E2E & 0.78 & 0.00 & 0.00 & 0.00 & 0.00 & 0.00 & 0.00 & 0.00 & 4.00 & 0.00 & 0.43 \\
NeSymReS & 4.65 & 1.29 & 0.00 & 0.00 & 4.63 & 0.00 & 5.56 & 0.00 & 12.00 & 6.67 & 15.84 \\
DSR & 0.00 & 0.00 & 0.00 & 0.00 & 0.00 & 0.00 & 0.00 & 0.00 & 0.00 & 0.00 & 1.66 \\
uDSR & 24.55 & 10.08 & 6.82 & 5.30 & 43.52 & 20.37 & 23.61 & 6.94 & 29.33 & 6.67 & 1.52 \\
PySR & 32.82 & 16.54 & 25.00 & 18.94 & 24.07 & 12.96 & 16.67 & 9.72 & 74.67 & 24.00 & \textbf{0.21} \\
\midrule
\multicolumn{12}{c}{\textit{Qwen3.6-35B-A3B}}\\
\midrule
LaSR & 10.08 & 4.65 & 6.82 & 4.55 & 8.33 & 0.00 & 4.17 & 4.17 & 24.00 & 12.00 & 25.61 \\
LLM-SR & 34.11 & 10.85 & 34.09 & 6.82 & 27.78 & 2.78 & 12.50 & 4.17 & 64.00 & 36.00 & 31.15 \\
SR-Scientist & 46.51 & 27.13 & 31.82 & 13.64 & 33.33 & 13.89 & 54.17 & 37.50 & 84.00 & 60.00 & 65.35 \\
HarnessSR & \textbf{65.37} & \textbf{38.76} & \textbf{53.03} & 32.58 & \textbf{61.11} & 22.22 & \textbf{68.06} & \textbf{41.67} & \textbf{90.67} & \textbf{70.67} & 3.66 \\
\midrule
\multicolumn{12}{c}{\textit{Qwen3-Coder-30B-A3B-Instruct}}\\
\midrule
LaSR & 8.79 & 6.46 & 4.55 & 3.79 & 12.04 & 9.26 & 5.56 & 5.56 & 14.67 & 8.00 & 9.60 \\
LLM-SR & 29.72 & 9.04 & 26.52 & 7.58 & 37.04 & 12.96 & 19.44 & 6.94 & 34.67 & 8.00 & 23.23 \\
SR-Scientist & 32.30 & 16.02 & 18.18 & 8.33 & 22.22 & 5.56 & 22.22 & 8.33 & 81.33 & 52.00 & 34.72 \\
HarnessSR & 57.36 & 37.73 & 37.12 & 33.33 & 58.33 & \textbf{28.70} & 61.11 & 27.78 & 88.00 & 68.00 & 4.03 \\
\midrule
\multicolumn{12}{c}{\textit{GPT-OSS-20B}}\\
\midrule
LaSR & 16.28 & 10.08 & 12.88 & 7.58 & 20.37 & 14.81 & 13.89 & 8.33 & 18.67 & 9.33 & 8.24 \\
LLM-SR & 28.94 & 10.85 & 18.18 & 3.79 & 24.07 & 4.63 & 18.06 & 2.78 & 65.33 & 40.00 & 39.58 \\
SR-Scientist & 48.06 & 25.32 & 35.61 & 16.67 & 54.63 & 25.93 & 40.28 & 22.22 & 68.00 & 42.67 & 15.37 \\
HarnessSR & 61.24 & 35.66 & 47.73 & \textbf{36.36} & 58.33 & 19.44 & 66.67 & \textbf{41.67} & 84.00 & 52.00 & 3.29 \\
\bottomrule
\end{tabular*}
\endgroup
\end{table}

\subsection{Experimental setup}\label{sec:experimental-setup}

\paragraph{Data.} Following SR-Scientist~\citep{xia2025sr}, we evaluate on LSR-Synth, the synthetic subset of LLM-SRBench~\citep{llmsrbench}, comprising 129 tasks across physics, chemistry, biology, and materials science. Harness optimization uses the 1,024 public synthetic training tasks released by SR-Scientist. The effect of training-set size is examined in Appendix~\ref{app:training-size}. Dataset details are provided in Appendix~\ref{app:training-data}.

\paragraph{Models and baselines.} We use Qwen3.6-35B-A3B, Qwen3-Coder-30B-A3B-Instruct, and GPT-OSS-20B. Qwen3.6-35B-A3B supports the main results and ablations; Qwen3-Coder-30B-A3B-Instruct supports the comparison with model-weight post-training; and GPT-OSS-20B tests applicability to another model family with fewer total parameters. The non-LLM baselines are GPLearn, E2E, NeSymReS, DSR, uDSR, and PySR. The LLM-based baselines are LaSR, LLM-SR, and SR-Scientist; SR-Scientist+RL is used for the post-training comparison. Model and search settings are given in Appendices~\ref{app:settings} and~\ref{app:harness}; baseline settings are in Appendix~\ref{app:baselines}.

\paragraph{Metrics.} Following LLM-SRBench~\citep{llmsrbench} and SR-Scientist~\citep{xia2025sr}, we use accuracy-to-tolerance, $\mathrm{Acc}_{\tau}$. For each task, we discard the 5\% of test points with the largest relative errors and check whether all remaining errors are at most $\tau$. We report the proportion of tasks meeting this criterion at $\tau=0.01$ and $0.001$, together with mean solving time per task. Evaluation and timing details are provided in Appendices~\ref{app:metrics} and~\ref{app:runtime}. We measure optimization effort by the total number of solver executions. Each execution is one run of an SR solver on one task and may include multiple LLM and SR-tool calls; repeated runs on the same task are counted separately. Symbolic-equivalence accuracy (SA) and supplementary OOD results are reported in Appendices~\ref{app:sa} and~\ref{app:ood}, respectively.

\subsection{Main results}\label{sec:main-results}

Table~\ref{tab:main} shows that HarnessSR's mean ID $\mathrm{Acc}_{0.01}$ and $\mathrm{Acc}_{0.001}$ across the three backbones are 45.0\% and 63.8\% higher than those of the next-best baseline, SR-Scientist, respectively. Mean solving time is reduced by 58.0--85.7\% relative to the fastest LLM-based baseline in each model group. Joint prompt--workflow optimization and SR-guided instruction refinement help translate solving feedback into more effective equation search.

These gains extend across scientific domains. HarnessSR leads at both accuracy thresholds in all four domains with the two Qwen models, and in physics, biology, and materials science with GPT-OSS-20B.

\subsection{Comparison with reinforcement-learning post-training}\label{sec:post-training-comparison}
\begin{table}[!htbp]
\centering
\caption{Two optimization routes based on Qwen3-Coder-30B-A3B-Instruct and their ID accuracy (\%).}
\label{tab:rl}
\small
\begin{tabular}{@{}llcc@{}}
\toprule
Method & Optimized component & ID $\mathrm{Acc}_{0.01}$ $\uparrow$ & ID $\mathrm{Acc}_{0.001}$ $\uparrow$\\
\midrule
SR-Scientist & None & 32.30 & 16.02\\
SR-Scientist+RL & Model weights & 40.92 & 20.69\\
HarnessSR & Prompts and workflow & \textbf{57.36} & \textbf{37.73}\\
\bottomrule
\end{tabular}
\end{table}

Table~\ref{tab:rl} compares weight post-training and harness optimization using the same Qwen3-Coder base model. HarnessSR achieves 57.36\% ID $\mathrm{Acc}_{0.01}$, compared with 40.92\% for SR-Scientist+RL, and also leads at ID $\mathrm{Acc}_{0.001}$. HarnessSR keeps model weights fixed and improves guidance for SR tools by optimizing role search instructions and their interactions. This result supports harness optimization as an effective route to improving symbolic regression performance. The one-time investments for both optimization routes are given in Appendix~\ref{app:cost}.

\subsection{LLM calls versus accuracy}\label{sec:llm-call-curves}
\begin{figure}[H]
\centering
\includegraphics[width=\linewidth]{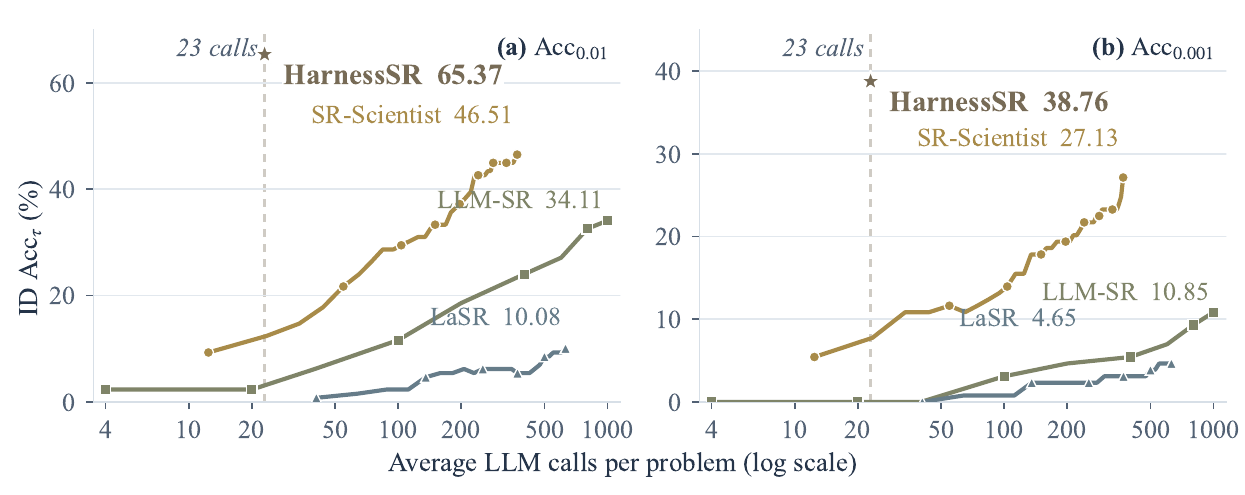}
\caption{Mean LLM calls per task versus ID accuracy on LSR-Synth (logarithmic horizontal axis). Baseline curves show call checkpoints; HarnessSR shows the full execution of its selected solver configuration. Curve construction is detailed in Appendix~\ref{app:call-checkpoints}.}
\label{fig:llm-calls}
\end{figure}

Figure~\ref{fig:llm-calls} compares accuracy and LLM calls using Qwen3.6-35B-A3B. HarnessSR achieves the highest accuracy at both thresholds with an average of 23 calls per task, about 6.2\% of the call count at SR-Scientist's final point. SR-Scientist, LLM-SR, and LaSR improve as calls accumulate, but their final accuracies remain below HarnessSR despite hundreds or up to a thousand calls. One-time optimization determines role search instructions, branch organization, and interaction rounds; the resulting configuration is then used directly on new tasks. The figure shows that reusing this configuration achieves higher accuracy with fewer model calls.

\subsection{Contribution of optimization stages}\label{sec:stage-ablation}
\begin{table}[!htbp]
\centering
\caption{ID accuracy (\%) and cumulative solver executions after each optimization stage (Qwen3.6-35B-A3B, LSR-Synth). The initial harness uses initial prompts and a single-branch workflow.}
\label{tab:stage}
\small
\begin{tabular}{@{}lccc@{}}
\toprule
Stage & ID $\mathrm{Acc}_{0.01}$ $\uparrow$ & ID $\mathrm{Acc}_{0.001}$ $\uparrow$ & Solver executions\\
\midrule
Initial harness & 5.43 & 1.55 & 0\\
After Stage 1 & 27.91 & 16.54 & 1,792\\
After Stage 2 & 43.41 & 32.82 & 2,752\\
After Stage 3 (full) & \textbf{65.37} & \textbf{38.76} & 4,728\\
\bottomrule
\end{tabular}
\end{table}

Table~\ref{tab:stage} shows that all three optimization stages improve accuracy. Stage 1 optimizes role prompts within basic workflows, establishing a useful starting point for the full solver. Stage 2 jointly searches workflows and prompt assignments, then briefly adapts role instructions to the selected workflows so that prompts fit their role combinations and interactions. ID $\mathrm{Acc}_{0.001}$ rises from 16.54\% to 32.82\%.

Stage 3 raises ID $\mathrm{Acc}_{0.01}$ from 43.41\% to 65.37\% and ID $\mathrm{Acc}_{0.001}$ from 32.82\% to 38.76\%. SR diagnostics connect candidate equations, residual patterns, and role actions, checking whether solving feedback leads to changes in search plans and whether different branches explore distinct equation structures. The reflection model uses these diagnostics to revise role instructions, improving guidance for subsequent equation search while keeping the workflow fixed.

\subsection{Workflow search and diagnostic feedback}\label{sec:workflow-feedback}
\begin{table}[!htbp]
\centering
\caption{Workflow search and feedback on LSR-Synth (Qwen3.6-35B-A3B): ID accuracy (\%) and mean LLM calls per task. Fixed-pool search selects presampled workflows; full-space search can propose new ones. Instructions are revised using execution traces directly or SR diagnostics of these traces.}
\label{tab:feedback}
\small
\begin{tabular}{@{}llccc@{}}
\toprule
Workflow search & Refinement feedback & ID $\mathrm{Acc}_{0.01}$ $\uparrow$ & ID $\mathrm{Acc}_{0.001}$ $\uparrow$ & LLM calls/task $\downarrow$\\
\midrule
Fixed-pool search & Execution traces & 49.35 & 28.17 & 37\\
Full-space search & Execution traces & 53.49 & 33.07 & \textbf{22}\\
Fixed-pool search & SR diagnostics & 56.59 & 32.30 & 37\\
Full-space search & SR diagnostics & \textbf{65.37} & \textbf{38.76} & 23\\
\bottomrule
\end{tabular}
\end{table}

Table~\ref{tab:feedback} shows that SR diagnostics improve accuracy at both thresholds under both search strategies, with larger gains in full-space search. The diagnostics connect candidate-equation structures and residual patterns with role actions, giving instruction revision concrete directions for equation search. This feedback guides role-prompt optimization more effectively than using execution traces directly, including when more workflows are available. Combining full-space search with SR diagnostics raises ID $\mathrm{Acc}_{0.01}$ from 49.35\% to 65.37\% and reduces mean LLM calls per task from 37 to 23, compared with fixed-pool search using execution traces.

\endgroup

\section{Conclusion and Limitation}
We introduced HarnessSR, a framework that treats the SR agent harness itself as an optimization target. It jointly optimizes role instructions, demonstrations, and workflow organization while keeping model weights and underlying search algorithms fixed. Results on LSR-Synth support this approach as a way to construct stronger, reusable SR agents. A central implication for SR is that numerical feedback can inform both equation refinement within a task and the development of search strategies reused across tasks. Through harness optimization, evidence from training tasks shapes how agents propose hypotheses, interpret residuals, and organize exploration and refinement on new problems. Prompts and workflows therefore encode consequential choices about how equation discovery proceeds. This perspective motivates studying solver organization as an optimization dimension in SR, alongside equation representations and search algorithms.

However, the present formulation searches within a predefined workflow space. Whether more open-ended workflow structures can reveal additional reusable strategies for equation discovery remains an important question for future research.

\bibliography{iclr2027_conference}
\bibliographystyle{iclr2027_conference}

\clearpage
\appendix
\begingroup
\normalem

\section{One-time Optimization Cost}\label{app:cost}
The main text compares two adaptation routes on the same Qwen3-Coder-30B-A3B-Instruct base model: SR-Scientist+RL updates model weights through reinforcement learning, whereas HarnessSR keeps the weights fixed and searches reusable role prompts and solver workflows. Table~\ref{tab:app-cost} reports the one-time investment required to obtain each final solver; the training task counts and solver execution counts refer to the optimization stage.

\begin{table}[!htbp]
\centering\small
\caption{One-time optimization investment for two adaptation routes.}\label{tab:app-cost}
\begin{tabularx}{\textwidth}{@{}lcc cX@{}}\toprule
Method & Training tasks & Solver executions & Optimization GPUs & Final product\\\midrule
SR-Scientist+RL & 1,024 & 15,360 & $32\times$ H200 & Updated model weights\\
HarnessSR & 160 & 4,728 & $2\times$ H100 & Role prompts and solver workflow\\\bottomrule
\end{tabularx}
\end{table}

Because of computational constraints, we did not reproduce the reinforcement learning training of SR-Scientist. Its evaluation results, training configuration, and model information are taken from the original paper, official code repository, and Hugging Face model page~\citep{xia2025sr}. Its 15,360 trajectories follow from 60 training steps, 32 prompts per step, and eight samples per prompt. Each sampled task-level trajectory counts as one solver execution in Table~\ref{tab:app-cost}. We deploy vLLM on two H100 GPUs to serve concurrent model requests during Harness search.

\section{Symbolic Accuracy}\label{app:sa}
Following the evaluation method of SR-Scientist~\citep{xia2025sr}, we use GPT-OSS-120B to assess the symbolic equivalence of predicted and ground-truth equations and report symbolic accuracy (SA).

\begin{table}[!htbp]
\centering
\caption{Symbolic accuracy (SA, \%) on LSR-Synth.}\label{tab:app-sa}
\small
\begin{minipage}[t]{0.22\textwidth}
\centering\footnotesize\textbf{Non-LLM methods}\par\smallskip
\begin{tabular*}{\linewidth}{@{\extracolsep{\fill}}lc@{}}\toprule
\textbf{Method} & \textbf{SA} $\uparrow$\\\midrule
uDSR & 0.52\\
PySR & 5.17\\\bottomrule
\end{tabular*}
\end{minipage}\hfill
\begin{minipage}[t]{0.75\textwidth}
\centering\footnotesize\textbf{LLM-based methods}\par\smallskip
\begin{tabular*}{\linewidth}{@{\extracolsep{\fill}}lccc@{}}\toprule
\textbf{Method} & \textbf{Qwen3.6} & \textbf{Qwen3-Coder} & \textbf{GPT-OSS}\\\midrule
LaSR & 6.46 & 2.84 & 5.68\\
LLM-SR & 5.43 & 4.91 & 4.65\\
SR-Scientist & 8.79 & 6.98 & 7.24\\
HarnessSR & \textbf{10.34} & \textbf{12.66} & \textbf{11.63}\\\bottomrule
\end{tabular*}
\end{minipage}
\par\smallskip
\parbox{\textwidth}{\footnotesize Qwen3.6: Qwen3.6-35B-A3B; Qwen3-Coder: Qwen3-Coder-30B-A3B-Instruct; GPT-OSS: GPT-OSS-20B.}
\end{table}

\section{Data, Evaluation, and Model Settings}\label{app:settings}
\subsection{Training Tasks and Scientific Domains}\label{app:training-data}
LLM-SRBench~\citep{llmsrbench} comprises LSR-Synth and LSR-Transform. LSR-Synth combines terms from established scientific equations with newly constructed terms, whereas LSR-Transform modifies existing equations. Following SR-Scientist~\citep{xia2025sr}, we use LSR-Synth because transformed equations may retain clues encountered during model pretraining. Each task provides observed data and separate in-distribution (ID) and out-of-distribution (OOD) test points.

Harness optimization uses a pool of 1,024 public synthetic training tasks released with SR-Scientist~\citep{xia2025sr}. The training-scale comparison uses 80, 160, 320, and 640 tasks. Training tasks provide role demonstrations, assess candidate prompts and workflows, and supply solver feedback for instruction revision. The final solver harness is frozen before evaluation on the 129 official LSR-Synth tasks.

LSR-Synth spans physics, chemistry, biology, and materials science. Table~\ref{tab:app-domains} summarizes representative tasks, input variables, and prediction targets in the four domains, following the examples in LLM-SRBench~\citep{llmsrbench}.

\begin{table}[!htbp]
\centering\small
\caption{Four scientific domains and evaluation task counts in LSR-Synth.}\label{tab:app-domains}
\begin{tabularx}{\textwidth}{@{}llXr@{}}\toprule
Domain & Typical problem & Observed variables and equation target & Tasks\\\midrule
Physics & Damped oscillators and other dynamics & Time $t$, position $x(t)$, velocity $v(t)$; acceleration relation & 44\\
Chemistry & Reaction kinetics & Time $t$, reactant concentration $c(t)$; rate of concentration change & 36\\
Biology & Population growth and dynamics & Time $t$, population $P(t)$; population growth rate & 24\\
Materials science & Stress--strain relations & Strain $\epsilon$ and temperature $T$; stress $\sigma$ & 25\\\midrule
Total & & & 129\\\bottomrule
\end{tabularx}
\end{table}

\subsection{Locked-Equation Evaluation and Aggregation}\label{app:metrics}
Each method outputs one locked equation per task. The external evaluator computes its predictions on separate in-distribution (ID) and out-of-distribution (OOD) test points; the solver receives observed data, variable descriptions, and scientific context. We use accuracy-to-tolerance, $\mathrm{Acc}_{\tau}$, following LLM-SRBench~\citep{llmsrbench} and SR-Scientist~\citep{xia2025sr}. For either distribution, let task $q$ have $n_q$ test points, with target $y_{q,i}$ and locked-equation prediction $\hat y_{q,i}$ at point $i$. For nonzero targets, the relative error is
\begin{equation}
e_{q,i}=\frac{|\hat y_{q,i}-y_{q,i}|}{|y_{q,i}|}.
\end{equation}
Sort the relative errors and discard the largest $\lfloor0.05n_q\rfloor$. Let $\mathcal I_q$ index the remaining test points. At threshold $\tau\in\{0.01,0.001\}$, the task-level indicator is
\begin{equation}
a_{q,\tau}=\mathbf{1}\!\left[\max_{i\in\mathcal I_q}e_{q,i}\leq\tau\right],
\end{equation}
where $\mathbf{1}[\cdot]$ equals one when its condition holds and zero otherwise. For a set of $N$ tasks, accuracy in percent is
\begin{equation}
\mathrm{Acc}_{\tau}=\frac{100}{N}\sum_{q=1}^{N}a_{q,\tau}.
\end{equation}
For domain $d$ with $n_d$ tasks, write $A_{d,\tau}=100\sum_{q\in d}a_{q,\tau}/n_d$. Overall accuracy weights domains by their task counts:
\begin{equation}
\mathrm{Acc}_{\tau}=\frac{44A_{\mathrm{physics},\tau}+36A_{\mathrm{chemistry},\tau}+24A_{\mathrm{biology},\tau}+25A_{\mathrm{materials},\tau}}{129}.
\end{equation}
ID and OOD evaluation use the same locked equation and numerical criterion, with test points drawn from different input distributions.

\subsection{Base Models and Per-Task Time}\label{app:runtime}
Qwen3.6-35B-A3B is used for the main results and method ablations. Qwen3-Coder-30B-A3B-Instruct supports the comparison with SR-Scientist+RL, which is post-trained from the same base model. GPT-OSS-20B shows the method's performance on another model family. For each base model, Harness optimization modifies role prompts and workflows while holding model weights fixed. The per-task Harness budget is at most 40 role-model calls and 24 symbolic regression (SR) backend calls. Both Qwen models are served through vLLM; the solver roles use LLM sampling temperature 0.7, while comparison methods follow their respective public decoding configurations.

Mean per-task time is measured from the start of solving a task to the locking of its final equation, including model calls and SR backend solving, and averaged over evaluation tasks. This metric measures the per-task running time of the optimized solver; the one-time Harness optimization cost is reported in Appendix~\ref{app:cost}.

For each base model, LaSR, LLM-SR, SR-Scientist, and HarnessSR are timed sequentially with exclusive use of the same GPU inference resources and serving configuration. The inference service uses vLLM 0.29.0, with six A100-SXM4-80GB GPUs allocated to each base model and two AMD EPYC 9654 CPUs. Each GPU hosts an independent inference instance, with a per-instance concurrent request limit of 64. For Qwen3-Coder, this limit is 27 due to key-value (KV) cache capacity constraints, and remains the same across methods using this model. Each method retains its own parallelism: LaSR issues up to 10 concurrent requests per task, and LLM-SR samples four candidates in parallel per prompt.

To check for resource contention during timing, we record running and queued request counts, KV cache occupancy, and CPU usage every 30 seconds. In the monitored samples, running request counts remain below the per-instance concurrency limits, no sustained queues are observed, and KV cache occupancy does not exceed 10\%. CPU utilization of the baseline jobs and the HarnessSR control process, normalized by their allocated core counts, does not exceed 60\%.

\subsection{Harness Optimization Settings}\label{app:optimization-settings}
The main experiment uses 160 training tasks divided into four mutually disjoint sets: $\mathcal D_1$, $\mathcal D_2$, and $\mathcal D_3$ each contain 32 tasks, and $\mathcal D_{\mathrm{eva}}$ contains 64 tasks. Stage $i$ uses $\mathcal D_i$ for optimization, while $\mathcal D_{\mathrm{eva}}$ is used for candidate selection.

Stage 1 performs six TPE configuration evaluations per role group. Stage 2 initializes $n=8$ workflows, comprising six basic configurations and two sampled configurations. Two prompt assignments per workflow produce 16 initial harnesses. TPE evaluates eight additional configurations, for 24 configurations in total. The top $K=6$ candidates undergo brief adaptation on an eight-task subset of $\mathcal D_2$, with two TPE configuration evaluations per candidate.

Stage 3 performs at most four instruction-refinement rounds for each of these six candidates. Prompt-bundle construction, candidate acceptance, and final selection follow the procedures in Section~\ref{sec:method}.

\section{Comparison Methods}\label{app:baselines}
The comparison methods span evolutionary search, neural-symbolic generation, LLM-assisted search, and autonomous tool interaction. We describe the mechanisms and experimental configurations of the ten baselines below.

\paragraph{GPLearn.} GPLearn evolves expression trees with genetic programming, improving fit across generations through selection, crossover, and mutation~\citep{gplearn}. The \texttt{gplearn} \texttt{Symbolic\allowbreak{}Regressor} uses 500 individuals, 200 generations, and tournament size 20. Its functions are addition, subtraction, multiplication, division, sine, cosine, square root, logarithm, absolute value, negation, and reciprocal. It provides a conventional evolutionary baseline without a pretrained model or LLM inference.

\paragraph{E2E.} E2E uses a pretrained Transformer to predict equation structure and constants directly from observations, followed by numerical refinement~\citep{e2e}. We use the public \texttt{facebookresearch/\allowbreak{}symbolicregression} implementation, at most 200 observed points per task, and refinement of 10 candidate expression trees. It tests whether pretrained end-to-end structure prediction can recover these new synthetic equations.

\paragraph{NeSymReS.} NeSymReS encodes observed points, generates symbolic expressions, and fits constants in candidate structures~\citep{nsr}. The public \texttt{Neural\allowbreak{}Symbolic\allowbreak{}Regression\allowbreak{}That\allowbreak{}Scales} implementation receives 500 points per task and uses beam width 32. It represents a neural generation approach with a distinct candidate search procedure from E2E.

\paragraph{DSR.} DSR samples expressions symbol by symbol with a recurrent neural network and uses risk-seeking policy gradients to increase the probability of high-scoring expressions~\citep{dsr}. The \texttt{deep-symbolic-optimization} implementation samples up to 100,000 expressions per task, with batch size 512 and learning rate 0.0005. Expression length ranges from 1 to 9; operators are addition, subtraction, multiplication, division, sine, cosine, exponential, and logarithm. Its task-specific learned search provides a direct reference for uDSR.

\paragraph{uDSR.} uDSR adds genetic programming and a polynomial \texttt{poly} operator to neural-guided expression search~\citep{landajuela2022unified}. Neural search retains DSR's sample budget, batch size, and learning rate. The genetic population is 100, run for 20 generations, with crossover and mutation probabilities of 0.5, tournament size 5, and maximum mutation-tree depth 3. Its operator set adds \texttt{poly} to DSR's operators. This comparison shows the effect of combining search mechanisms.

\paragraph{PySR.} PySR searches equations through multiple evolutionary populations and maintains a trade-off between fit and expression complexity~\citep{pysr}. The independent baseline runs one backend solve per task with all input variables and operators: 125 iterations, 550 cycles per iteration, 15 populations of size 33, maximum expression size 40, and randomization weight 0.1. Binary operators are addition, subtraction, multiplication, division, and power; unary operators are exponential, logarithm, square root, sine, and cosine. Its final equation selection rule is given in Appendix~\ref{app:backend-pysr}. This baseline uses the same class of SR backend as HarnessSR, making the contribution of role guidance and workflow coordination visible.

\paragraph{LaSR.} LaSR maintains an LLM-proposed natural-language concept library within evolutionary equation search and uses those concepts to generate and modify equations~\citep{lasr}. We use \texttt{Library\allowbreak{}Augmented\allowbreak{}Symbolic\allowbreak{}Regression.jl}: 25 iterations, 550 cycles each, 10 populations of size 33, maximum expression size 30, and the concept library enabled. The LLM operation weights for mutation, crossover, and random generation are all 0.001. Operators and nesting constraints match the independent PySR baseline. The response limit is 4,096 tokens, with up to 1,000 model requests per task.

\paragraph{LLM-SR.} LLM-SR represents equations as programs, uses scientific context and strong existing candidates to propose new programs, and advances the search through execution and constant fitting~\citep{llmsr}. We use the baseline configuration released with LLM-SRBench~\citep{llmsrbench}, allowing at most 1,000 model calls per task. Each prompt samples four independent candidates, with one call counted per candidate. The method's candidate selector determines the final equation and selects from candidates available at each call checkpoint in Figure~\ref{fig:llm-calls}.

\paragraph{SR-Scientist.} SR-Scientist uses an LLM to call data-analysis and equation-evaluation tools, revises hypotheses from observed feedback, and carries a cache of previous strong equations across iterations~\citep{xia2025sr}. We use the public inference framework, with at most 40 iterations and 25 interaction rounds per iteration; the cache provides the top three equations to the next iteration. Temperature is 0.7, \texttt{top\_p=1.0}, the response limit is 8,192 tokens, and the initial mean absolute percentage error target is 0.1\%. The evaluator fits constants with BFGS. Its public candidate selection rule determines the final equation. Iterations, interaction rounds, and the cache govern different steps; Figure~\ref{fig:llm-calls} uses actual call counts to describe its inference demand.

\paragraph{SR-Scientist+RL.} This variant post-trains Qwen3-Coder-30B-A3B-Instruct through reinforcement learning and retains SR-Scientist's inference-time tool workflow. We use the released \texttt{GAIR/SR-Scientist-30B} model~\citep{xia2025sr}; its one-time training investment is given in Appendix~\ref{app:cost}. It compares model-weight updates with Harness optimization in final solving performance.

\subsection{Call-Accuracy Trajectories}\label{app:call-checkpoints}
Figure~\ref{fig:llm-calls} uses the same 129 evaluation tasks. The twelve LLM-SR checkpoints take the first 4, 8, 12, 16, 20, 40, 100, 200, 400, 600, 800, or 1,000 LLM calls for each task. Its original selection rule based on observed data locks the equation available by that checkpoint. Table~\ref{tab:app-llmsr-curve} reports each point's accuracy. The checkpoint call count is the actual cumulative calls made by the method on each task.

\begin{table}[!htbp]
\centering\small
\caption{LLM-SR call checkpoints and ID accuracy over 129 tasks (\%).}\label{tab:app-llmsr-curve}
\begin{tabular}{@{}rcc@{}}\toprule
Cumulative calls & ID $\mathrm{Acc}_{0.01}$ $\uparrow$ & ID $\mathrm{Acc}_{0.001}$ $\uparrow$\\\midrule
4 & 2.33 & 0.00\\8 & 2.33 & 0.00\\12 & 2.33 & 0.00\\16 & 2.33 & 0.00\\20 & 2.33 & 0.00\\
40 & 6.20 & 0.00\\100 & 11.63 & 3.10\\200 & 18.60 & 4.65\\400 & 24.03 & 5.43\\
600 & 27.13 & 6.98\\800 & 32.56 & 9.30\\1,000 & 34.11 & 10.85\\\bottomrule
\end{tabular}
\end{table}

At the end of SR-Scientist iteration $r$, its selection rule based on observed data chooses one equation from each task's candidates through that iteration. Tasks that end early retain their final equation and cumulative call count in later iterations. The horizontal coordinate is the mean of cumulative calls on the 129 tasks, $x_r=129^{-1}\sum_{q=1}^{129}c_{q,r}$, where $c_{q,r}$ is the number of calls actually made for task $q$ through iteration $r$; the vertical coordinate is overall accuracy at that iteration. Table~\ref{tab:app-srs-curve} lists iterations 1, 5, 10, 15, 20, 25, 30, 35, and 40. The HarnessSR point is the aggregate after the frozen solver completes on the same task set.

\begin{table}[!htbp]
\centering\small
\caption{SR-Scientist iteration checkpoints and ID accuracy over 129 tasks (\%).}\label{tab:app-srs-curve}
\begin{tabular}{@{}rccc@{}}\toprule
Iteration & Mean cumulative calls & ID $\mathrm{Acc}_{0.01}$ $\uparrow$ & ID $\mathrm{Acc}_{0.001}$ $\uparrow$\\\midrule
1 & 12.40 & 9.30 & 5.43\\5 & 54.69 & 21.71 & 11.63\\10 & 103.37 & 29.46 & 13.95\\
15 & 149.98 & 33.33 & 17.83\\20 & 196.41 & 37.21 & 19.38\\25 & 240.91 & 42.64 & 21.71\\
30 & 284.14 & 44.96 & 22.48\\35 & 328.45 & 44.96 & 23.26\\40 & 369.84 & 46.51 & 27.13\\\bottomrule
\end{tabular}
\end{table}

LaSR checkpoint $r$ takes the candidates after the $10r$-th population return for each task, $r=1,\ldots,25$. Figure~\ref{fig:llm-calls} plots all 25 rounds and a separately locked run-completion point. Each checkpoint uses LaSR's selection rule based on observed data before evaluation by the same external evaluator. The horizontal coordinate averages the actual cumulative LLM calls over 129 tasks, and the vertical coordinate uses all 129 tasks as its denominator. Table~\ref{tab:app-lasr-curve} reports rounds 1, 5, 10, 15, 20, and 25, and run completion. Model calls continue after round 25, so the final cumulative count is slightly higher while accuracy is unchanged.

\begin{table}[!htbp]
\centering\small
\caption{LaSR iteration checkpoints, actual mean cumulative calls, and ID accuracy over 129 tasks (\%).}\label{tab:app-lasr-curve}
\begin{tabular}{@{}lccc@{}}\toprule
Iteration & Mean cumulative calls & ID $\mathrm{Acc}_{0.01}$ $\uparrow$ & ID $\mathrm{Acc}_{0.001}$ $\uparrow$\\\midrule
1 & 40.63 & 0.78 & 0.00\\5 & 135.33 & 4.65 & 2.33\\10 & 253.52 & 6.20 & 2.33\\
15 & 371.05 & 5.43 & 3.10\\20 & 499.16 & 8.53 & 3.88\\25 & 627.83 & 10.08 & 4.65\\
Run completion & 629.21 & 10.08 & 4.65\\\bottomrule
\end{tabular}
\end{table}

\section{Symbolic Regression Backends and Candidate Selection}\label{app:harness} With continued advances in the field, symbolic regression (SR) has attracted increasing attention~\citep{chen2022symbolic,xu2023retention,meidani2023snip,zhang2025rag,li2023discrete,e2e,li2024focus,xia2025sr,llmsr}. Beyond numerical prediction, SR transforms observations into testable scientific hypotheses, enabling equation discovery across physics, biology, materials science, and other domains~\citep{yu2025discover,li2023adaptive,hu2024domain,hu2026dass,hu2025context,li2026score,hu2026beyond}. Among the diverse families of SR methods, we select two representative approaches—a GP-based solver and an LLM-based solver—as the base SR tools in HarnessSR. By covering fundamentally different search paradigms, this choice demonstrates that HarnessSR can integrate heterogeneous SR algorithms as underlying tools rather than being restricted to a specific class of solvers.

We compare three SR backends: PySR, LLM equation generation, and a hybrid of LLM and PySR. For each backend, we independently run all three stages of harness optimization to search for its role prompts and solver workflow. Each resulting harness is then frozen for evaluation.

\subsection{PySR Backend}\label{app:backend-pysr}
The main experiments use the PySR backend, with the same base settings as the independent PySR baseline in Appendix~\ref{app:baselines}. Roles can specify subsets of input variables and operators, structural constraints, and initial expressions; the adapter maps this guidance into PySR search settings. The independent PySR baseline runs one backend solve with the full variable and operator sets.

The selector computes candidate relative root mean square error on observed points with nonzero targets. Let $e^*$ be the lowest error. It first retains candidates with error no greater than $e^*+\max(0.05e^*,10^{-4})$ and then chooses the expression with the fewest nodes. Thus, when fits are close, the simpler expression is preferred. HarnessSR and the independent PySR baseline use this same final selection rule.

\subsection{LLM Equation-Generation Backend}\label{app:backend-llm}
For an LLM equation generator, the adapter includes search guidance and structural hypotheses in the equation-generation prompt and turns expression guesses into initial program skeletons with free parameters. The generator then creates candidates, fits constants, and evaluates them. This backend draws on LLM-SR's equation-program generation and constant fitting~\citep{llmsr}, and on SR-Scientist's use of execution feedback to improve candidates~\citep{xia2025sr}. Its input includes scientific context, variable descriptions, a summary of observed data, and the roles' structural hypotheses, analysis, and repair directions. The equation generator and solver roles use the same base model. Its role prompts and solver workflow are obtained through independent harness optimization with the LLM equation-generation backend.

One backend call consists of two rounds of generation and evaluation. Following LLM-SR's sampling design, each round independently samples eight responses, and each response may provide up to ten candidate programs; one backend call therefore makes at most sixteen generation requests. Candidate programs are parsed, checked for numerical validity, and deduplicated by equation skeleton. Free constants are fitted with Levenberg--Marquardt nonlinear least squares from up to 32 starting points. Fitted candidates are scored on observed data; stronger equations and their scores are retained as generation experience. The second round uses the first round's fitting feedback to generate and evaluate further candidates. Equations and evaluations from both rounds are returned to the Harness for subsequent diagnosis, repair, and final equation selection.

\subsection{Hybrid Backend}\label{app:backend-hybrid}
The hybrid configuration independently searches for its own role prompts and solver workflow. For a given problem, it executes complete solves with the LLM equation-generation backend and the PySR backend, pools the candidates from both routes, and selects one final equation. Both the LLM-only and hybrid configurations evaluate candidates using the 95th percentile of relative error on observed points, together with expression complexity and numerical validity; simpler expressions are preferred when errors are close. The merge makes no further LLM calls or constant fits.

\begin{table}[!htbp]
\centering\small
\caption{ID accuracy (\%) on LSR-Synth with LLM-based and hybrid SR backends for Qwen3.6-35B-A3B. Role prompts and workflows are independently optimized for each backend.}\label{tab:app-backends}
\begin{tabular}{@{}lcc@{}}\toprule
SR backend & ID $\mathrm{Acc}_{0.01}$ $\uparrow$ & ID $\mathrm{Acc}_{0.001}$ $\uparrow$\\\midrule
LLM & 59.43 & 34.11\\
LLM + PySR & \textbf{66.67} & \textbf{40.05}\\\bottomrule
\end{tabular}
\end{table}

Table~\ref{tab:app-backends} shows that the hybrid configuration achieves higher accuracy than the LLM-only configuration and slightly exceeds the PySR configuration used in the main experiments. These results support the applicability of harness optimization to different equation-search backends.

\section{Out-of-Distribution Results}\label{app:ood}
Figures~\ref{fig:app-ood-qwen}--\ref{fig:app-ood-gptoss} compare ID and OOD accuracy by base model; Table~\ref{tab:app-ood} reports the OOD results.

\begin{figure}[H]
\centering
\includegraphics[width=\linewidth]{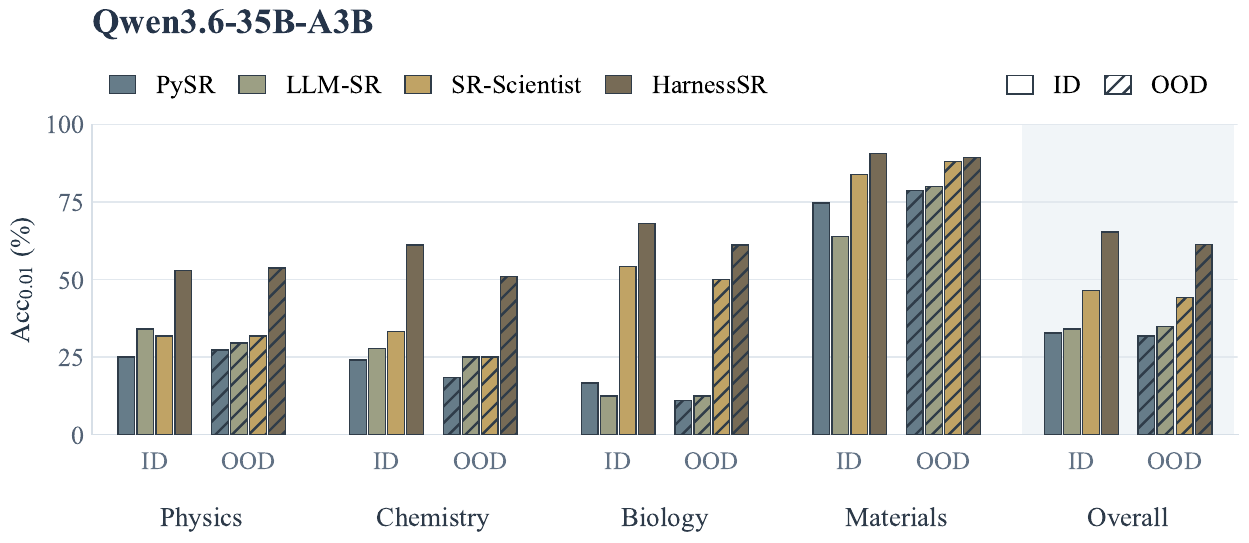}
\caption{ID and OOD $\mathrm{Acc}_{0.01}$ (\%) for methods using Qwen3.6-35B-A3B.}\label{fig:app-ood-qwen}
\end{figure}

\begin{figure}[H]
\centering
\includegraphics[width=\linewidth]{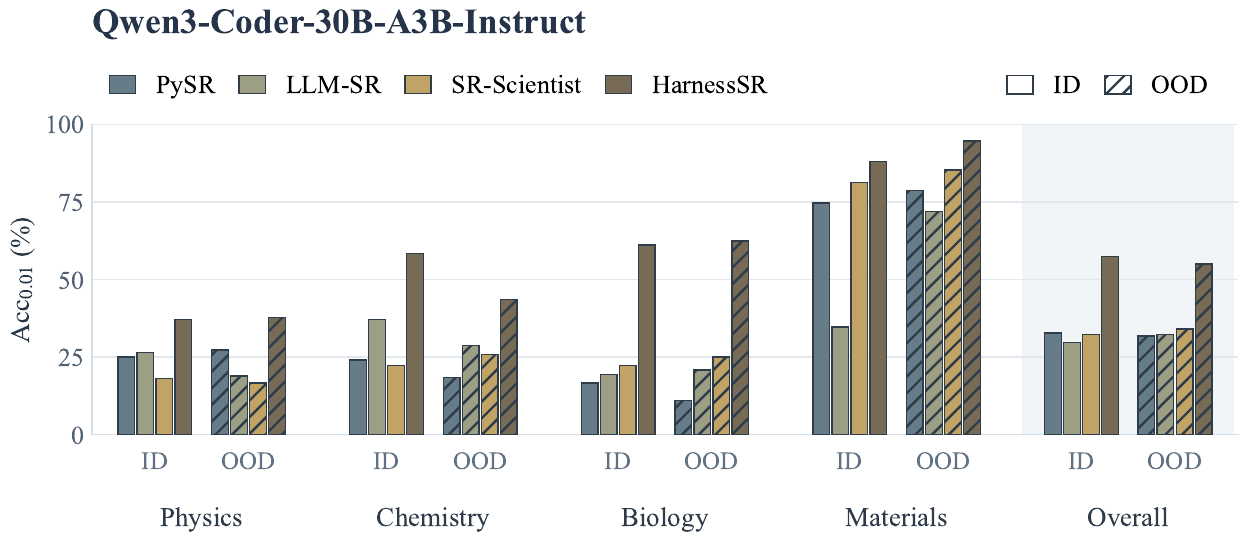}
\caption{ID and OOD $\mathrm{Acc}_{0.01}$ (\%) for methods using Qwen3-Coder-30B-A3B-Instruct.}\label{fig:app-ood-coder}
\end{figure}

\begin{figure}[H]
\centering
\includegraphics[width=\linewidth]{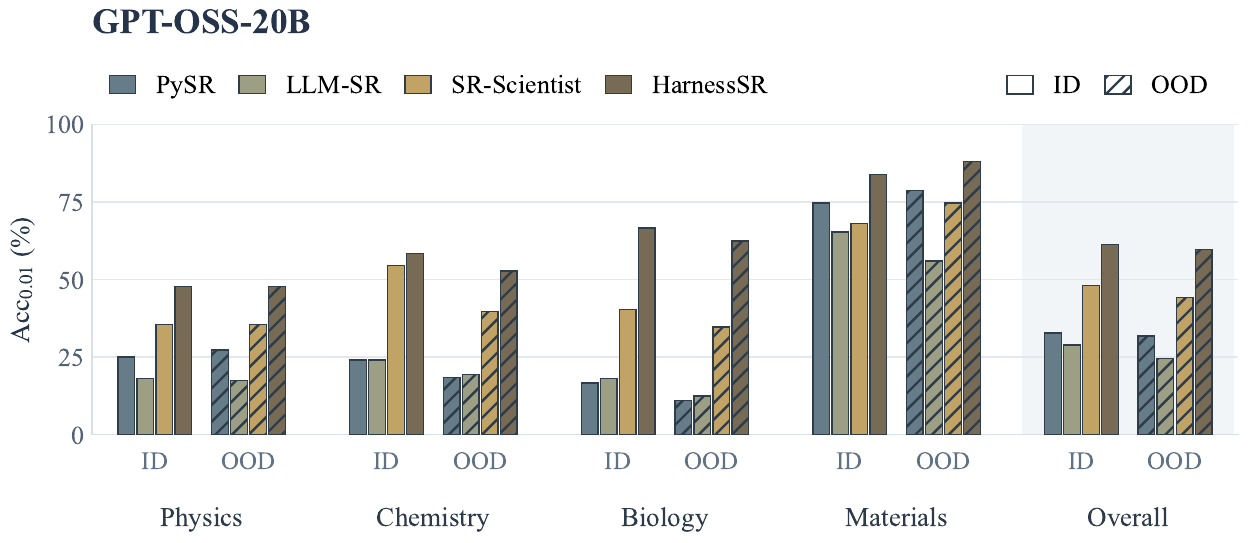}
\caption{ID and OOD $\mathrm{Acc}_{0.01}$ (\%) for methods using GPT-OSS-20B.}\label{fig:app-ood-gptoss}
\end{figure}

\begin{table}[!htbp]
\centering\small
\caption{OOD accuracy (\%) on LSR-Synth.}\label{tab:app-ood}
\setlength{\tabcolsep}{4pt}
\begin{tabular*}{\textwidth}{@{\extracolsep{\fill}}lccccc@{}}\toprule
\multirow{2}{*}{\textbf{Method}} & \textbf{Overall} & \textbf{Physics} & \textbf{Chemistry} & \textbf{Biology} & \textbf{Materials}\\
& $\mathrm{Acc}_{0.01}$ $\uparrow$ & $\mathrm{Acc}_{0.01}$ $\uparrow$ & $\mathrm{Acc}_{0.01}$ $\uparrow$ & $\mathrm{Acc}_{0.01}$ $\uparrow$ & $\mathrm{Acc}_{0.01}$ $\uparrow$\\\midrule
\multicolumn{6}{c}{\textit{Non-LLM methods}}\\\midrule
PySR & 31.78 & 27.27 & 18.52 & 11.11 & 78.67\\\midrule
\multicolumn{6}{c}{\textit{Qwen3.6-35B-A3B}}\\\midrule
LLM-SR & 34.88 & 29.55 & 25.00 & 12.50 & 80.00\\
SR-Scientist & 44.19 & 31.82 & 25.00 & 50.00 & 88.00\\
HarnessSR & \textbf{61.24} & \textbf{53.79} & \textbf{50.93} & \textbf{61.11} & \textbf{89.33}\\\midrule
\multicolumn{6}{c}{\textit{Qwen3-Coder-30B-A3B-Instruct}}\\\midrule
LLM-SR & 32.30 & 18.94 & 28.70 & 20.83 & 72.00\\
SR-Scientist & 34.11 & 16.67 & 25.93 & 25.00 & 85.33\\
HarnessSR & \textbf{55.04} & \textbf{37.88} & \textbf{43.52} & \textbf{62.50} & \textbf{94.67}\\\midrule
\multicolumn{6}{c}{\textit{GPT-OSS-20B}}\\\midrule
LLM-SR & 24.55 & 17.42 & 19.44 & 12.50 & 56.00\\
SR-Scientist & 44.19 & 35.61 & 39.81 & 34.72 & 74.67\\
HarnessSR & \textbf{59.69} & \textbf{47.73} & \textbf{52.78} & \textbf{62.50} & \textbf{88.00}\\\bottomrule
\end{tabular*}
\end{table}

On Qwen3.6-35B-A3B, HarnessSR reaches 61.24\% OOD $\mathrm{Acc}_{0.01}$, above LLM-SR (34.88\%) and SR-Scientist (44.19\%) on the same base model. Across physics, chemistry, biology, and materials science, its accuracy is 53.79\%, 50.93\%, 61.11\%, and 89.33\%, compared with 31.82\%, 25.00\%, 50.00\%, and 88.00\% for SR-Scientist. On Qwen3-Coder-30B-A3B-Instruct, HarnessSR reaches 55.04\% overall, with 37.88\%, 43.52\%, 62.50\%, and 94.67\% across the four domains.

Compared with the main-text ID results, HarnessSR changes from 65.37\% to 61.24\% OOD $\mathrm{Acc}_{0.01}$ on Qwen3.6, a difference of 4.13 percentage points; on Qwen3-Coder-30B-A3B-Instruct it changes from 57.36\% to 55.04\%, a difference of 2.32 points. HarnessSR reaches 59.69\% OOD $\mathrm{Acc}_{0.01}$ on GPT-OSS-20B.

\section{Solver-Role Instructions}\label{app:prompts}
\subsection{Initial Instructions for Five Roles}
The following five instructions are starting points for role-prompt search. Search generates multiple candidate instructions and combines them with examples to select prompt packages. At each call, task context, observations, and search state are supplied as runtime input; the symbolic regression (SR) backend executes the search plans. References to PySR in the original instructions are retained.

\paragraph{Initial Planner.} The Planner generates an SR search plan from scientific context and observations.
\begin{tcblisting}{breakable,enhanced,title after break={Initial Planner (continued)},colback=promptfill,colframe=promptstroke,boxrule=0.6pt,arc=2pt,boxsep=3pt,left=3pt,right=3pt,top=3pt,bottom=3pt,listing only,listing options={basicstyle=\ttfamily\footnotesize,breaklines=true,columns=fullflexible,keepspaces=true,showstringspaces=false,breakatwhitespace=false}}
Propose a PySR search plan as JSON: binary_operators, unary_operators,
feature_subset, complexity, constraints, guesses. No compute/budget fields.
\end{tcblisting}

\paragraph{Initial Analyst.} The Analyst organizes observations and scientific context into notes for subsequent search.
\begin{tcblisting}{breakable,enhanced,title after break={Initial Analyst (continued)},colback=promptfill,colframe=promptstroke,boxrule=0.6pt,arc=2pt,boxsep=3pt,left=3pt,right=3pt,top=3pt,bottom=3pt,listing only,listing options={basicstyle=\ttfamily\footnotesize,breaklines=true,columns=fullflexible,keepspaces=true,showstringspaces=false,breakatwhitespace=false}}
Analyze the observed data and scientific context; produce notes that guide
the symbolic search. Never sees Truth or held-out rows.
\end{tcblisting}

\paragraph{Initial Diagnoser.} The Diagnoser uses residuals of the current candidate to identify missing structure and choose the next-round policy.
\begin{tcblisting}{breakable,enhanced,title after break={Initial Diagnoser (continued)},colback=promptfill,colframe=promptstroke,boxrule=0.6pt,arc=2pt,boxsep=3pt,left=3pt,right=3pt,top=3pt,bottom=3pt,listing only,listing options={basicstyle=\ttfamily\footnotesize,breaklines=true,columns=fullflexible,keepspaces=true,showstringspaces=false,breakatwhitespace=false}}
Diagnose why the current best candidate is still wrong, from its observed
residuals, and decide how the next round should act: EXPLORE a different
functional form, or REPAIR the current one.
\end{tcblisting}

\paragraph{Initial Repairer.} The Repairer proposes the next search plan from the diagnosis and policy.
\begin{tcblisting}{breakable,enhanced,title after break={Initial Repairer (continued)},colback=promptfill,colframe=promptstroke,boxrule=0.6pt,arc=2pt,boxsep=3pt,left=3pt,right=3pt,top=3pt,bottom=3pt,listing only,listing options={basicstyle=\ttfamily\footnotesize,breaklines=true,columns=fullflexible,keepspaces=true,showstringspaces=false,breakatwhitespace=false}}
Act on the diagnosis under the given policy and emit the next PySR search
plan as JSON. Under 'repair' keep the current structure and fix it; under
'explore' propose a different functional family. A debate round additionally
supplies the other branches' best candidates as peer_candidates.
\end{tcblisting}

\paragraph{Initial Synthesizer.} The Synthesizer combines branch candidates into a plan for further search.
\begin{tcblisting}{breakable,enhanced,title after break={Initial Synthesizer (continued)},colback=promptfill,colframe=promptstroke,boxrule=0.6pt,arc=2pt,boxsep=3pt,left=3pt,right=3pt,top=3pt,bottom=3pt,listing only,listing options={basicstyle=\ttfamily\footnotesize,breaklines=true,columns=fullflexible,keepspaces=true,showstringspaces=false,breakatwhitespace=false}}
Synthesize across all branches' candidates into one refined PySR search
plan (JSON).
\end{tcblisting}

\subsection{Final PySR Solver Workflow and Role Instructions}
This section presents the final solver workflow and role instructions used with the PySR backend in the main experiments. The solve begins with Analyst examining the task context, variables, and observed data. Three parallel search branches then run: in each branch, Planner prepares an initial plan and invokes the SR backend, followed by three rounds of Diagnoser diagnosis, Repairer plan revision, and backend solving. Synthesizer collects candidate equations across branches, proposes an additional plan, and runs the backend once more. All candidates from initial search, repair, and synthesis are evaluated and selected together to produce the final equation. Full execution uses 23 role-model calls and 13 SR-backend calls.

\begin{figure}[H]
\centering
\includegraphics[width=\linewidth]{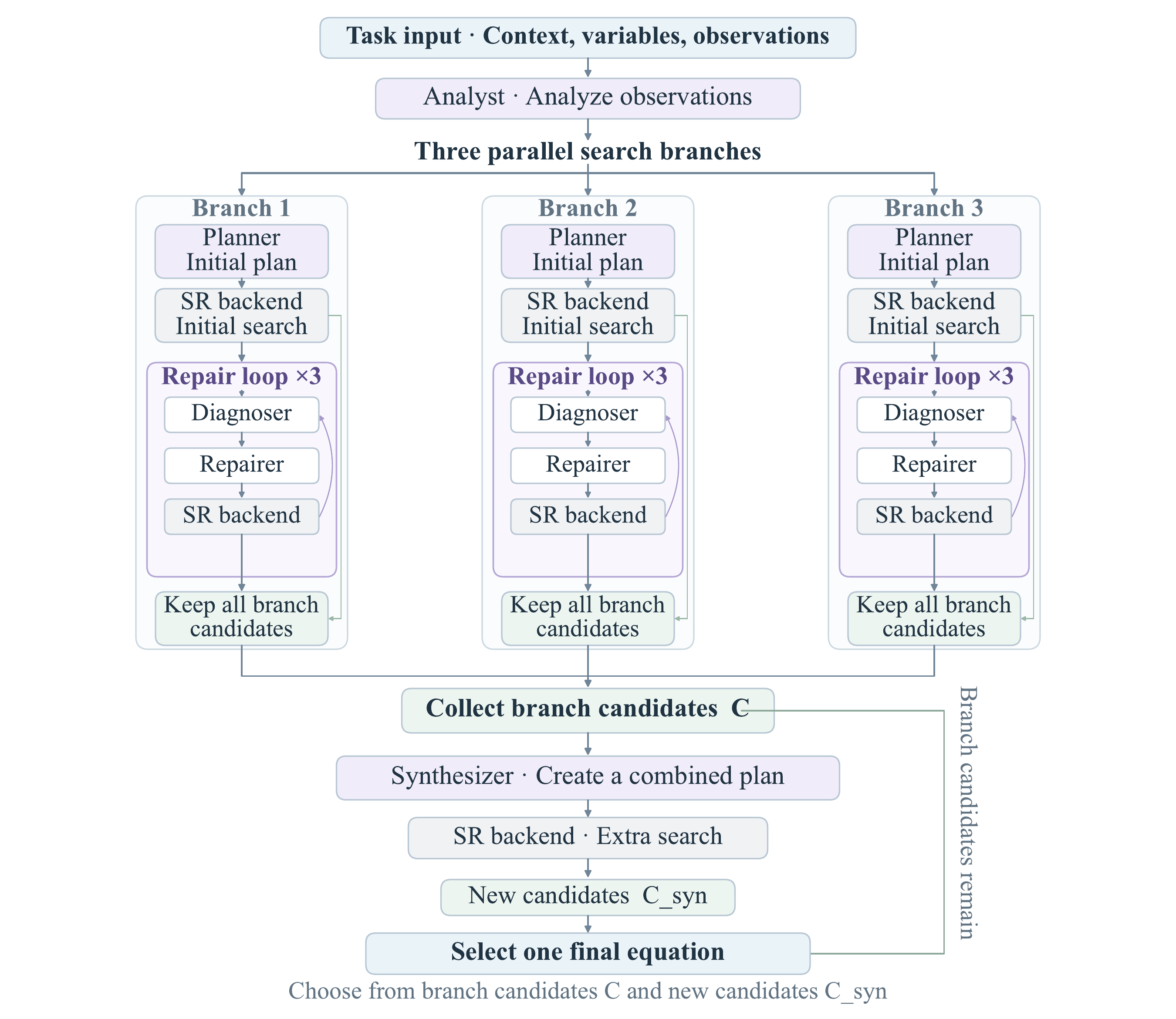}
\caption{Final solver workflow for the PySR backend in the main experiments. Three branches search and repair independently; after synthesis, the final equation is selected from all candidates.}\label{fig:app-selected-solver}
\end{figure}

\paragraph{Analyst instruction.}
\begin{tcblisting}{breakable,enhanced,title after break={Analyst instruction (continued)},colback=promptfill,colframe=promptstroke,boxrule=0.6pt,arc=2pt,boxsep=3pt,left=3pt,right=3pt,top=3pt,bottom=3pt,listing only,listing options={basicstyle=\ttfamily\footnotesize,breaklines=true,columns=fullflexible,keepspaces=true,showstringspaces=false,breakatwhitespace=false}}
You are a scientific data analyst specializing in symbolic regression. Your task is to examine the provided scientific context, variable descriptions, and deterministic observed data evidence.

Based on this information, generate concise analysis notes that will guide a symbolic search algorithm to discover the underlying mathematical equation. Your notes should:
1. Identify the likely type of relationship (e.g., linear, exponential, polynomial, trigonometric, differential).
2. Highlight key physical or chemical constraints implied by the variable names and context.
3. Note any apparent symmetries, invariants, or dominant terms.
4. Suggest potential functional forms or variable combinations to explore.

**Crucial Constraints:**
- You must NOT use any ground truth equations or held-out data rows.
- Base your analysis strictly on the provided context and the visible data patterns.
- Keep the output focused and actionable for a symbolic regression engine.
\end{tcblisting}

\paragraph{Planner instruction.}
\begin{tcblisting}{breakable,enhanced,title after break={Planner instruction (continued)},colback=promptfill,colframe=promptstroke,boxrule=0.6pt,arc=2pt,boxsep=3pt,left=3pt,right=3pt,top=3pt,bottom=3pt,listing only,listing options={basicstyle=\ttfamily\footnotesize,breaklines=true,columns=fullflexible,keepspaces=true,showstringspaces=false,breakatwhitespace=false}}
You are an expert scientific modeling architect tasked with designing a search strategy for the PySR symbolic regression engine. Your goal is to propose a precise, constrained, and effective search plan that will guide the engine in discovering the underlying mathematical law governing the observed data.

Based on the provided **scientific context** (including domain knowledge, variable definitions, and observed data evidence), and any optional **analysis notes** or **perspective hints**, generate a JSON object representing the search plan.

Adhere strictly to the following guidelines for constructing the plan:

1.  **Operators**:
    *   **`binary_operators`**: Select from `["+", "-", "*", "/", "^"]`. Prioritize operations that are physically plausible for the domain (e.g., avoid division if variables can be zero unless physically justified).
    *   **`unary_operators`**: Select from `["+", "-", "sin", "cos", "tan", "exp", "log", "sqrt", "abs"]`. Use these to capture common physical relationships (e.g., logarithmic scaling, exponential decay/growth, periodicity).

2.  **Feature Subset**:
    *   **`feature_subset`**: Specify the list of input feature indices (0-based) to be used. If no specific constraints exist, use all available features (`[0, 1, ..., N-1]`). If certain features are known to be irrelevant or noisy, exclude them.

3.  **Complexity**:
    *   **`complexity`**: Define the complexity constraints for the generated expressions. This should be a dictionary or list indicating limits on expression size, depth, or number of operations. Keep it reasonable to balance model fit and interpretability.

4.  **Constraints**:
    *   **`constraints`**: List any logical, physical, or mathematical constraints the resulting formula must satisfy (e.g., "positive only", "periodic with period 2*pi", "asymptotic behavior"). If none, provide an empty list.

5.  **Guesses**:
    *   **`guesses`**: Provide a list of initial candidate expressions (as strings) to seed the search. These should be informed by the context and perspective. Examples: `"x0 * x1"`, `"sin(x0) + x1"`, `"exp(-x0) * x1"`. If no strong prior exists, provide a few simple combinations.

**Output Format**:
Return a single valid JSON object with the following keys: `binary_operators`, `unary_operators`, `feature_subset`, `complexity`, `constraints`, and `guesses`. Do not include any other text or markdown formatting outside the JSON block. Do not include compute or budget fields.
\end{tcblisting}

\paragraph{Diagnoser instruction.}
\begin{tcblisting}{breakable,enhanced,title after break={Diagnoser instruction (continued)},colback=promptfill,colframe=promptstroke,boxrule=0.6pt,arc=2pt,boxsep=3pt,left=3pt,right=3pt,top=3pt,bottom=3pt,listing only,listing options={basicstyle=\ttfamily\footnotesize,breaklines=true,columns=fullflexible,keepspaces=true,showstringspaces=false,breakatwhitespace=false}}
Analyze the residuals of the current best candidate equation to diagnose why it fails to fit the data. Based on the structure of the errors, determine if the current functional form is fundamentally flawed or just needs parameter adjustments. Output a diagnosis of the missing or incorrect structural elements, and select a policy: "explore" to search for a different functional family, or "repair" to fix the current one.
\end{tcblisting}

\paragraph{Repairer instruction.}
\begin{tcblisting}{breakable,enhanced,title after break={Repairer instruction (continued)},colback=promptfill,colframe=promptstroke,boxrule=0.6pt,arc=2pt,boxsep=3pt,left=3pt,right=3pt,top=3pt,bottom=3pt,listing only,listing options={basicstyle=\ttfamily\footnotesize,breaklines=true,columns=fullflexible,keepspaces=true,showstringspaces=false,breakatwhitespace=false}}
You are an expert Symbolic Regression Strategist. Your goal is to generate a precise JSON search plan for the PySR engine to improve the fit of a differential equation model.

**Input Context:**
1.  **Task Context**: The scientific domain, variables, and evidence. Note the mapping of input variable names (e.g., `epsilon`, `T`) to PySR placeholders (`x1`, `x2`).
2.  **Diagnosis**: A detailed explanation of why the current best candidate failed (residual analysis).
3.  **Policy**: Either `'repair'` (fix the current structure) or `'explore'` (try a new functional family).
4.  **Candidates**: Current expressions and their errors.
5.  **Peer Candidates**: (Optional) Best candidates from other branches, provided if this is a "debate" round.
6.  **Perspective**: (Optional) A hint about the desired functional form.

**Instructions:**

1.  **Analyze the Diagnosis & Policy**:
    *   If **Policy is 'repair'**: Identify the specific structural flaw indicated by the diagnosis (e.g., missing non-linearity, incorrect variable interaction, wrong time-dependency). You **MUST** maintain the general functional form family (e.g., if the current best is polynomial, stay polynomial; if trigonometric, stay trigonometric) but modify terms, coefficients, or operations to address the error. Do not switch families unless the diagnosis explicitly states the family is fundamentally wrong and the policy allows for a radical shift (rare in 'repair').
    *   If **Policy is 'explore'**: The current family is fundamentally insufficient. Propose a completely different functional family (e.g., switching from polynomial to trigonometric, exponential, or rational functions) that better captures the underlying dynamics suggested by the residuals.

2.  **Leverage Peer Context (if available)**:
    *   If `peer_candidates` are provided, consider their successes. If they fit well, analyze *why* (what features do they have?) and ensure your new plan incorporates similar strengths if exploring, or differentiates if repairing to avoid local minima.

3.  **Construct the Search Plan**:
    *   Output a valid JSON object.
    *   The JSON must include a `description` field summarizing your strategic change. In the description, explicitly map the input variable names (e.g., `epsilon`, `T`) to the PySR placeholders (`x1`, `x2`) you are using.
    *   The JSON must include a `constraints` field (optional) and an `operators` field.
        *   **Operators**: Only include valid PySR unary and binary operators. Valid unary operators: `sin`, `cos`, `tan`, `exp`, `log`, `sqrt`, `abs`, `neg`. Valid binary operators: `+`, `-`, `*`, `/`, `pow` (do NOT use `^`).
        *   **Constraints**: Only include valid PySR constraint keys with integer values, such as `max_depth`, `max_nodes`, `complexity`. Do NOT invent constraint names like `force_variable_inclusion` or `max_power`.
    *   Ensure the plan is executable by PySR (e.g., valid syntax for operations).
    *   If `guesses` are provided, use `x1` and `x2` placeholders consistent with the input context mapping.

**Output Format:**
Return **only** the JSON object. Do not include markdown formatting, backticks, or explanatory text outside the JSON.

**Example Structure:**
```json
{
  "description": "Switching from polynomial to trigonometric family to capture oscillatory residuals. Note: epsilon maps to x1, T maps to x2.",
  "operators": {
    "unary": ["sin", "cos", "exp", "log", "sqrt"],
    "binary": ["+", "-", "*", "/", "pow"]
  },
  "constraints": {
    "max_depth": 10,
    "max_nodes": 20
  },
  "variables": ["x1", "x2"],
  "complexity_limit": 15
}
```

**Key Constraints:**
*   Be specific in your `description`.
*   Ensure `operators` are compatible with the chosen `policy` and are valid PySR operators.
*   If `policy` is 'repair', do not change the core functional family entirely unless the diagnosis explicitly demands a radical shift; instead, tweak terms.
*   If `policy` is 'explore', propose a distinct family.
*   Never use `^` as an operator; use `pow` if exponentiation is needed.
*   Never use invalid constraint names like `max_power`, `max_degree`, or `force_variable_inclusion`.
\end{tcblisting}

\paragraph{Synthesizer instruction.}
\begin{tcblisting}{breakable,enhanced,title after break={Synthesizer instruction (continued)},colback=promptfill,colframe=promptstroke,boxrule=0.6pt,arc=2pt,boxsep=3pt,left=3pt,right=3pt,top=3pt,bottom=3pt,listing only,listing options={basicstyle=\ttfamily\footnotesize,breaklines=true,columns=fullflexible,keepspaces=true,showstringspaces=false,breakatwhitespace=false}}
Synthesize across all branches' candidates into one refined PySR search
plan (JSON).
\end{tcblisting}

\section{Successful Equation-Recovery Cases}\label{app:cases}
We select three successful cases involving population growth, a nonlinear oscillator, and material stress to show the recovered equation structures and their ID and OOD predictions. ID and OOD denote the two test distributions, and NMSE is normalized mean squared error. A ``Pass'' in Table~\ref{tab:app-cases} means that the task meets the indicated accuracy threshold on that distribution. Each case has 500 ID and 500 OOD points.

\begin{table}[!htbp]
\centering\small
\caption{Numerical results for three successful equation-recovery cases.}\label{tab:app-cases}
\begin{tabular*}{\textwidth}{@{\extracolsep{\fill}}lcccc@{}}\toprule
Case (domain) & ID NMSE & OOD NMSE & \shortstack{$\mathrm{Acc}_{0.01}$\\ID / OOD} & \shortstack{$\mathrm{Acc}_{0.001}$\\ID / OOD}\\\midrule
BPG10 (population growth) & $2.9237\times10^{-15}$ & $4.5305\times10^{-14}$ & Pass/Pass & Pass/Pass\\
PO18 (physical oscillator) & $1.4546\times10^{-15}$ & $1.6621\times10^{-15}$ & Pass/Pass & Pass/Pass\\
MatSci22 (materials science) & $1.2594\times10^{-11}$ & $2.4181\times10^{-10}$ & Pass/Pass & Pass/Pass\\\bottomrule
\end{tabular*}
\end{table}

\subsection{Population Growth: BPG10}
This task takes time $t$ and population size $P$ as inputs and predicts the population growth rate. The ground truth and locked prediction are
\begin{align}
\frac{\mathrm dP}{\mathrm dt}
&\approx 0.1010081003\,P^{0.3333333333}
       +0.1010081003\,P,\\
\widehat{\frac{\mathrm dP}{\mathrm dt}}
&=0.1010081\,P^{0.3333332}
       +0.1010081\,P.
\end{align}
Both expressions combine a fractional-power term and a linear term under a shared coefficient; time does not appear explicitly in the final equation. At smaller $P$, the $P^{1/3}$ term is relatively more prominent, and its contribution relative to the linear term changes approximately as $P^{-2/3}$ as $P$ grows. The prediction preserves this joint nonlinear and linear dependence, with fitted exponent and coefficients close to the ground truth. The two predicted terms have the same coefficient at the recorded precision, matching the ground truth's shared-coefficient form. ID and OOD NMSE are $2.9237\times10^{-15}$ and $4.5305\times10^{-14}$, respectively, and the equation passes both accuracy thresholds on both sets of test points.

\subsection{Nonlinear Oscillator: PO18}
The inputs are position $x$, time $t$, and velocity $v$; the target is the derivative of velocity. The ground truth combines a linear displacement-restoring term with nonlinear velocity coupling. The locked prediction retains the same variable combinations:
\begin{align}
\frac{\mathrm dv}{\mathrm dt}
&=-\omega_0^2x-\beta v\sin(x),\\
\widehat{\frac{\mathrm dv}{\mathrm dt}}
&=-1.8126475\,x-0.583656\,v\sin(x).
\end{align}
The ground-truth $\beta$ and $\omega_0$ remain the dataset's symbolic parameters, while the prediction gives fitted numerical coefficients. The $x$ term changes oppositely with position, and $v\sin(x)$ couples velocity to position. Since $\sin(x)$ changes sign, the direction of this coupling varies with position. Both terms vanish at $x=0$; away from that point, position affects both the restoring term and the magnitude and direction of velocity coupling. The prediction adds no explicit time term and retains the ground truth's concise dependence on state variables. ID and OOD NMSE are both below $2\times10^{-15}$, and the equation passes both thresholds.

\subsection{Material Stress: MatSci22}
This task predicts stress $\sigma$ from strain $\epsilon$ and temperature $T$. The ground truth combines a quadratic strain term with a linear strain term that varies sinusoidally with temperature; the prediction uses cosine to express the same periodic family:
\begin{align}
\sigma
&\approx 72.33109008\,\epsilon^2
 +7.74654609\,\epsilon\sin(T-287.77390919),\\
\hat\sigma
&=\epsilon\bigl[72.33109\,\epsilon
 -7.746545\cos(T-3.4597688)\bigr].
\end{align}
Using $-\cos z=\sin(z-\pi/2)$ and sine periodicity, the predicted periodic term can be written as $7.746545\,\epsilon\sin(T-287.77390395)$. Thus both expressions combine $\epsilon^2$ and $\epsilon\sin(T-\phi)$. Their phases differ by approximately $5.24\times10^{-6}$ radians, and the coefficients differ only slightly through numerical fitting. ID and OOD NMSE are $1.2594\times10^{-11}$ and $2.4181\times10^{-10}$, respectively, with passes at both accuracy thresholds. The case shows how the same periodic structure can be expressed using sine or cosine and a phase shift.

\section{Effect of Training-Set Size}\label{app:training-size}\label{sec:training-size}
\begin{table}[!htbp]
\centering
\caption{ID accuracy (\%) and cumulative solver executions across training-set sizes (Qwen3.6-35B-A3B, LSR-Synth). Each nested training set undergoes all three optimization stages; 160 tasks correspond to the main setting.}
\label{tab:size}
\small
\begin{tabular}{@{}cccc@{}}
\toprule
Training tasks & ID $\mathrm{Acc}_{0.01}$ $\uparrow$ & ID $\mathrm{Acc}_{0.001}$ $\uparrow$ & Solver executions\\
\midrule
80 & 52.71 & 28.94 & 2,648\\
160 & 65.37 & 38.76 & 4,728\\
320 & 68.48 & 41.34 & 9,016\\
640 & 69.51 & 42.12 & 16,312\\
\bottomrule
\end{tabular}
\end{table}

Table~\ref{tab:size} shows that expanding the training set consistently improves the optimized solver's accuracy. From 80 to 160 tasks, $\mathrm{Acc}_{0.01}$ rises from 52.71\% to 65.37\%, with improved accuracy at the stricter threshold as well. Accuracy continues to improve at 320 and 640 tasks, though each doubling brings a smaller gain. More training tasks provide richer solving feedback for role-prompt and workflow optimization, helping the resulting configuration recover equations on new tasks.

Further accuracy gains require greater one-time search effort. Relative to the main 160-task setting, 640 tasks entail about 3.45 times as many solver executions, raise $\mathrm{Acc}_{0.01}$ from 65.37\% to 69.51\%, and further improve accuracy at the stricter threshold. The 160-task setting achieves substantial accuracy with moderate search effort, while larger training sets leave room for further gains. The results connect training scale, solver quality, and upfront search effort in harness optimization.

\endgroup

\end{document}